\documentclass{article}
\usepackage{ijcai26}

\usepackage{times}
\usepackage[hidelinks]{hyperref}
\usepackage[utf8]{inputenc}
\usepackage{graphicx}
\usepackage{subcaption}
\usepackage{amsmath}
\usepackage{booktabs}
\usepackage{natbib}
\usepackage{amssymb}
\usepackage[switch]{lineno}
\usepackage{forest}
\usetikzlibrary{arrows.meta,shapes,positioning,trees}

\title{STELLAR: Spatio-Temporal Environmental Learning with Latent Alignment and Refinement for Long-Tailed Species Distribution Modeling}
\author{
    Shufeng Kong\textsuperscript{1,2}, Tao Yu\textsuperscript{1}, Yuanyuan Wei\textsuperscript{1}, Caihua Liu\textsuperscript{5,3,2}\thanks{Correspondance to Caihua Liu: cl2869@cornell.edu}, Junwen Bai\textsuperscript{2}, Yingheng Wang\textsuperscript{2}, Marc Grimson\textsuperscript{2}, Daniel Fink\textsuperscript{4}, Carla P. Gomes\textsuperscript{2}
    \affiliations
    \textsuperscript{1} School of Software Engineering, Sun Yat-sen University, Zhuhai, China \\
    \textsuperscript{2} Department of Computer Science, Cornell University, Ithaca, NY, USA \\
    \textsuperscript{3} Department of Ecology and Evolutionary Biology, Cornell University, Ithaca, NY, USA \\
    \textsuperscript{4} Cornell Lab of Ornithology, Ithaca, NY, USA \\
    \textsuperscript{5} School of Computer Science and Artificial Intelligence, Foshan University, Foshan, China
    \emails
    \{sk2299, cl2869, jb2467, yw2349, mg2425, daniel.fink\}@cornell.edu, gomes@cs.cornell.edu, \{yutao67,weiyy53\}@mail2.sysu.edu.cn
}

\begin{document}

\maketitle

\begin{abstract}
Joint Species Distribution Modeling (JSDM) is a key enabler for biodiversity monitoring and conservation planning. However, accurate JSDM faces two coupled challenges: environmental drivers and species distributions are inherently spatio-temporal, while species co-occurrence patterns exhibit complex non-linear community structure and severe long-tail imbalance driven by rare species. Existing approaches often address these factors in isolation, learning from static covariates or neglecting the historical trajectories of dynamic community structure. To overcome these limitations, we propose \textbf{STELLAR} (\textbf{S}patio-\textbf{T}emporal \textbf{E}nvironmental \textbf{L}earning with \textbf{L}atent \textbf{A}lignment and \textbf{R}efinement), a novel framework that learns a shared latent space where dynamic habitat context and community structure are optimized jointly. Our approach integrates three complementary components: (1) a Graph-Temporal Encoder that employs graph attention and recurrent units to aggregate spatial neighborhood effects and capture the co-evolving historical dynamics of environmental context and community structure; (2) a Context-Anchored Latent Alignment mechanism that structures the latent space using a label-activated mixture prior and supervised contrastive learning, actively clustering species based on shared environmental preferences; and (3) an Imbalance-Aware Decoupled Decoding module that utilizes Asymmetric Loss to focus learning on hard, rare species samples, preventing mode collapse in the long tail. Experiments on the large-scale eBird dataset, curated with domain experts, demonstrate that our framework significantly outperforms state-of-the-art baselines, particularly in predicting rare species and revealing interpretable species interactions.
\end{abstract}

\section{Introduction}

The accelerating biodiversity crisis, exacerbated by anthropogenic climate change and habitat fragmentation, necessitates scalable computational paradigms to monitor ecosystems, anticipate range shifts, and prioritize conservation interventions \citep{gomes2019computational}. While Species Distribution Models (SDMs) \citep{elith2009species} serve as the cornerstone of this effort, linking occurrences to environmental covariates, ecological communities are not merely aggregates of independent species. Real-world assemblages are governed by two tightly coupled forces: (i) \textit{extrinsic spatio-temporal dynamics}, where habitat suitability is shaped by accumulating environmental conditions and historical trajectories, and (ii) \textit{intrinsic biotic structure}, which induces complex, non-linear co-occurrence regimes and severe long-tail imbalance, where the majority of species are rare but ecologically critical \citep{wisz2013role, ovaskainen2011making}.
Despite advances in deep learning for ecology, current Joint Species Distribution Models (JSDMs) often confront a fundamental mismatch between this ecological reality and modeling practice. Here we identify three critical gaps in the existing literature:

\paragraph{The Spatio-Temporal Representation Gap.}
On the \textit{input side}, standard pipelines typically treat location-time samples as independent, static snapshots. This assumption violates two key ecological regularities: \textit{spatial spillover} (source-sink dynamics) and \textit{temporal persistence} (ecological memory). 
While statistical methods like Gaussian Processes (GPs) \citep{golding2016fast} and recent Spatial Implicit Neural Representations (SINR) \citep{cole2023spatial} improve spatial continuity, they generally assume isotropic smoothing. They fail to model \textit{anisotropic barriers} (e.g., rivers blocking dispersal) and rarely capture the causal neighborhood graph explicitly. Furthermore, autoregressive models used in weather forecasting \citep{fan2022gnn} have yet to be effectively adapted to the irregular, sparse nature of biodiversity survey data, leaving the co-evolution of environments and assemblages under-modeled.

\paragraph{The Structural Alignment Gap.}
On the \textit{output side}, community prediction requires learning high-dimensional label structures. Classical statistical approaches, such as Hierarchical Modeling of Species Communities (HMSC) \citep{ovaskainen2017make} and Deep Multivariate Probit (DMVP) \citep{chen2018end}, rely on Multivariate Probit (MVP) formulations. While statistically rigorous, MVP models scale cubically with species richness $\mathcal{O}(S^3)$, rendering them intractable for large-scale species monitoring.
Deep latent-variable approaches, such as Multivariate Probit Variational AutoEncoder (MPVAE) \citep{bai2020disentangled}, offer scalable alternatives. However, these models typically rely on restrictive unimodal Gaussian priors ($\mathcal{N}(0,I)$). In heterogeneous ecosystems, this isotropic assumption forces distinct community types (e.g., forest guilds vs. grassland guilds) to collapse into a single mode. While Contrastive Gaussian Mixture VAEs (C-GMVAE) \citep{bai2022cgmvae} have attempted to introduce multimodal priors, they rely on static feature encoders, failing to align the latent community structure with the dynamic trajectory of the environmental context and species assemblages.

%\paragraph{The Long-Tail Inference Gap.}
%Finally, a critical failure mode in JSDM is the handling of rare species. Biodiversity data follows a heavy-tailed distribution; even in species-rich areas, most species are absent in any given survey. Standard reconstruction objectives, such as Binary Cross Entropy (BCE), are dominated by these abundant ``easy negatives.'' While computer vision has addressed imbalance via Focal Loss or Logit Adjustment \citep{golding2016fast}, these techniques are difficult to apply in generative, multi-label settings where resampling is impossible (as one sample contains both head and tail species). Consequently, generative JSDMs frequently collapse into a trivial degenerate solution: predicting ``absent'' for all rare species, rendering the model ineffective for conservation planning.

\paragraph{The Long-Tail Inference Gap.}
Finally, a critical failure mode in JSDM is the handling of rare species. Biodiversity data follows a heavy-tailed distribution; even in species-rich areas, most species are absent in any given survey. Standard reconstruction objectives, such as Binary Cross Entropy (BCE), are dominated by these abundant ``easy negatives.'' While computer vision has successfully addressed imbalance via loss re-weighting strategies like {Focal Loss} \citep{lin2017focal} or {Logit Adjustment} \citep{menon2020long}, these advances have not been widely adopted in generative ecological modeling. Furthermore, traditional {data resampling} techniques are effectively impossible in multi-label settings, as a single survey frequently contains both head (common) and tail (rare) species. Consequently, generative JSDMs, such as C-GMVAE and MPVAE, frequently collapse into a trivial degenerate solution: predicting ``absent'' for all rare species, rendering the model ineffective for conservation planning.

\paragraph{Stakeholder Collaboration.} 
This work was conducted in direct collaboration with domain experts and conservation biologists from the Cornell Lab of Ornithology. Their roles included: 
(1) \textbf{Problem Formulation:} Our partners identified the failure of current models to predict rare species (``the long tail") as a primary bottleneck for conservation planning, directly motivating our Imbalance-Aware design; 
(2) \textbf{Data Curation:} Domain experts facilitated access to quality-controlled detection pipelines and advised on the selection of ecologically relevant environmental covariates; and 
(3) \textbf{Validation:} Biologists qualitatively evaluated the ecological validity of our model predictions and latent structure. This ensures the proposed framework meets the operational standards required for biodiversity monitoring.

\paragraph{Our Contribution.}
In this work, we address these challenges by reformulating JSDM as a {joint representation learning and robust structural alignment problem}. We propose a novel framework that synthesizes the co-evolving trajectories of environmental context and species assemblages with an imbalance-aware generative model. Our framework integrates three tightly coupled components:

%First, we learn {spatio-temporal habitat embeddings} with a combined Graph Neural Network (GNN) and Recurrent Neural Network (RNN) encoder that (i) propagates information across spatial neighborhoods and (ii) summarizes historical trajectories of both covariates and species assemblages into context-rich representations. This produces a dynamic spatio-temporal embedding that encapsulates both spatial dependence and temporal evolution, serving as a critical ecological context signal for downstream community modeling.

First, we learn {spatio-temporal habitat embeddings} via a hybrid {Graph-Temporal Encoder}. By integrating Graph Neural Networks (GNN) with Recurrent Neural Networks (RNN), this module (i) propagates information across anisotropic spatial neighborhoods to capture dispersal effects, and (ii) summarizes the historical trajectories of both environmental covariates and antecedent community states. This process yields a context-rich, dynamic embedding that encapsulates the co-evolution of space and time, serving as a critical conditioning signal for downstream community modeling.

%Second, we adopt a {Context-Anchored Latent Alignment} mechanism to resolve multimodal community structures. We formulate a {label-activated mixture prior} where species prototypes act as learnable centroids conditioned on the spatio-temporal embedding. Leveraging supervised contrastive learning, this mechanism creates a dynamic ``gravitational field'' in the latent space: embeddings of co-occurring species are actively pulled toward their shared habitat trajectory.

Second, we adopt a {Context-Anchored Latent Alignment} mechanism to resolve multimodal community structures. We formulate a {label-activated mixture prior} utilizing learnable species prototypes as distinct cluster centroids. By leveraging supervised contrastive learning, this mechanism aligns the {spatio-temporal habitat embedding} with these centroids, creating a dynamic ``gravitational field'' in the latent space: the environmental representation of a site is actively pulled toward the prototypes of its co-occurring species, ensuring the learned context captures the correct ecological regime.

%Third, to robustly model the ``long tail'' of biodiversity, we adopt an {Imbalance-Aware Decoding} strategy. Instead of standard reconstruction, we integrate an Asymmetric Loss (ASL) \citep{ridnik2021asymmetric} directly into the variational objective. This induces a ``hard-mining'' effect during generation, dynamically down-weighting the gradients from abundant easy negatives to focus model capacity on rare, difficult-to-predict species.

Third, to robustly model the ``long tail'' of biodiversity, we introduce an {Imbalance-Aware Decoupled Decoding} strategy. By integrating {Asymmetric Loss (ASL)} \citep{ridnik2021asymmetric} directly into the variational objective, we replace standard reconstruction with a dynamic focusing mechanism. This induces a ``hard-mining'' effect during generation, systematically down-weighting the gradients from abundant easy negatives to redirect model capacity toward the rare, difficult-to-predict species that drive conservation priorities.

%Our contributions are summarized as follows:
%\begin{itemize}
%    \item \textbf{Unified Spatio-Temporal JSDM Framework.} We propose a novel architecture that synthesizes graph-based environmental dynamics and generative structure learning into a single end-to-end pipeline, moving beyond static snapshots to capture the evolving ecological context.
%    \item \textbf{Context-Anchored Latent Alignment.} We employ a conditional generative framework utilizing a label-activated mixture prior. By integrating supervised contrastive learning, this mechanism actively clusters co-occurring species into distinct latent modes, effectively capturing the multimodality of community structures.
%    \item \textbf{Imbalance-Aware Generative Reasoning.} We explicitly tackle the long-tail problem by reformulating the variational objective with an Asymmetric Loss strategy. This prevents mode collapse in rare species prediction, a persistent issue in standard generative JSDMs.
%\end{itemize}

\begin{figure*}[ht]
    \centering
    \includegraphics[width=0.92\linewidth]{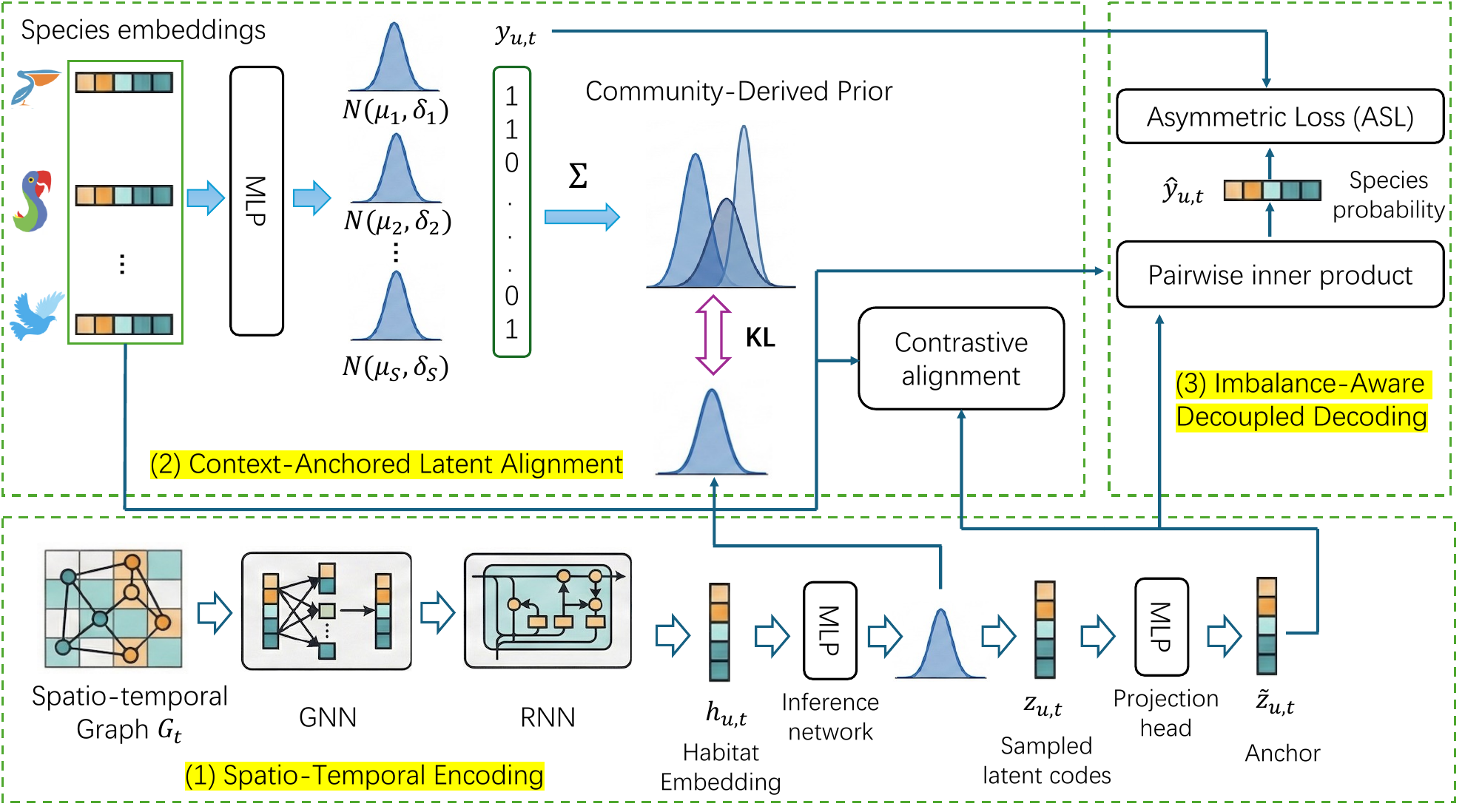}
    \caption{The architeture of our model STELLAR: (1) A Spatio-Temporal Encoder (GNN+RNN) fuses regional and historical cues into a habitat embedding $\mathbf{h}_{u,t}$. (2) The Context-Anchored Latent Alignment module aligns the variational posterior with a label-activated Gaussian Mixture Prior using KL divergence and contrastive learning. Note that while the distributions are depicted as univariate contours for visual clarity, they are modeled as high-dimensional multivariate Gaussians in the latent space. (3) An Imbalance-Aware Decoupled Decoding module utilizes ASL to focus learning on hard, rare species samples, preventing mode collapse in the long tail.} \label{fig:arch}
\end{figure*}

\section{Methodology}

Accurate biodiversity forecasting requires resolving two coupled challenges: modeling the \textit{extrinsic spatio-temporal dynamics} of the environment, where drivers propagate across locations and accumulate over history, and capturing the \textit{intrinsic multimodal structure} of species assemblages, which exhibit severe class imbalance. 
We propose a novel framework comprising three components: (1) a \textbf{GNN-RNN Encoder} for context-rich biophysical embedding; (2) a \textbf{Context-Anchored Latent Alignment} mechanism for multimodal structure learning; and (3) an \textbf{Imbalance-Aware Decoupled Decoding} module for robust long-tail prediction. Figure~\ref{fig:arch} illustrates the archtecture of our model.

\subsection{Preliminaries and Problem Formulation}

Let $\mathcal{S} = \{(u, t, \mathbf{x}_{u,t}, \mathbf{y}_{u,t})\}$ denote a set of observed survey records, where $u$ represents a specific geographic location and $t$ is a discrete time step. For each record, we observe a local environmental covariate vector $\mathbf{x}_{u,t} \in \mathbb{R}^{D}$ (e.g., land cover) and a multi-label species presence-absence vector $\mathbf{y}_{u,t} \in \{0, 1\}^{S}$, where $S$ is the species richness.

To explicitly model regional connectivity and historical dynamics despite irregular sampling, we discretize the spatial domain into a grid system $\mathcal{C} = \{c_1, \dots, c_M\}$. We construct a dynamic spatial graph $\mathcal{G}_t = (\mathcal{C}, \mathcal{E}_t)$, where nodes $c \in \mathcal{C}$ represent grid cells and edges $e \in \mathcal{E}_t$ encode spatial adjacency or functional connectivity between cells.

We define the \textit{regional context} for a grid cell $c$ as a temporal window of size $W$ containing aggregated covariates and community states: $\mathbf{H}^{\text{grid}}_{c,t} = [(\bar{\mathbf{x}}_{c, \tau}, \bar{\mathbf{y}}_{c, \tau})]_{\tau=t-W}^{t-1}$, where $\bar{\mathbf{x}}$ and $\bar{\mathbf{y}}$ are pooled averages of all surveys within cell $c$ at time $\tau$.

Our objective is to learn a probabilistic model $P_\Theta(\mathbf{y}_{u,t} | \mathbf{x}_{u,t}, \mathbf{H}^{\text{grid}}_{c(u),t}, \mathcal{G})$ that maximizes the likelihood of the observed assemblages, conditioned on both the high-resolution local environment $\mathbf{x}_{u,t}$ and the coarse-grained regional history $\mathbf{H}^{\text{grid}}_{c(u),t}$ of the enclosing cell $c(u)$.

\iffalse
\subsection{Preliminaries and Problem Formulation}
Let $\mathcal{G} = (\mathcal{V}, \mathcal{E})$ denote a spatial graph where nodes $u \in \mathcal{V}$ ($|\mathcal{V}|=N$) represent survey locations and edges $e \in \mathcal{E}$ encode spatial adjacency.
For each location $u$ at discrete time step $t$, we observe a vector of environmental covariates $\mathbf{x}_{u,t} \in \mathbb{R}^{D}$ (e.g., temperature, precipitation, topography) and a multi-label species presence-absence vector $\mathbf{y}_{u,t} \in \{0, 1\}^{S}$, where $S$ is the species richness. 

To capture historical environmental effects and species community dynamics, we define the input context as a temporal window of covariates $\mathbf{X}_{u,t} = [\mathbf{x}_{u, t-W+1}, \dots, \mathbf{x}_{u,t}]$ and past species states $\mathbf{Y}^{\text{hist}}_{u,t} = [\mathbf{y}_{u, t-W+1}, \dots, \mathbf{y}_{u,t-1}]$. Note that we explicitly include past species states in the input to model {persistence} and {priority effects}.

Our objective is to learn a probabilistic model $P_\Theta(\mathbf{y}_{u,t} | \mathbf{X}_t, \mathbf{Y}^{\text{hist}}_t, \mathcal{G})$ that maximizes the likelihood of the observed community assemblages across all locations and time steps.
\fi

\subsection{Spatio-Temporal Habitat Representation}

Ecological dynamics are governed by processes that are both spatially diffusive (e.g., dispersion) and temporally cumulative (e.g., extinction debts). Standard Convolutional Neural Networks (CNNs) assume a fixed grid structure, which is often mismatched with the irregular distribution of ecological survey data. To capture these dynamics while retaining local precision, we introduce a hierarchical \textbf{Point-Grid Fusion strategy} backed by a serialized {GNN-RNN architecture}.

\paragraph{Spatio-Temporal Grid Aggregation.}
Ecological survey data is frequently unstructured and opportunistic (e.g., citizen science checklists), leading to irregular spatial sampling. To regularize this input, we discretize the spatial domain into a grid system. For each time step $t$, we map all available survey checklists to their corresponding grid cells. 

For a grid cell $c$, we compute an aggregated feature vector $\mathbf{x}_{c,t}$ by pooling the environmental covariates (averaged within the cell) and the species presence-absence records (aggregated from all checklists in the cell). This aggregation mitigates sampling noise and aligns disparate data sources into a structured representation.

\paragraph{Grid-Based Graph Construction.}
We construct the spatial graph $\mathcal{G}_t = (\mathcal{C}, \mathcal{E}_t)$ where nodes $c \in \mathcal{C}$ represent the grid cells. Edges $e \in \mathcal{E}_t$ are defined based on spatial adjacency (connecting neighboring grid cells) or strictly defined distance thresholds between grid centroids. This topology allows the subsequent GNN layers to propagate information across the grid lattice, effectively modeling the diffusion of species and environmental drivers from neighboring regions into the focal cell.

It is worth noting that while our current grid adjacency approach captures dominant spatial gradients, the framework is extensible. Future iterations could define edges using functional connectivity matrices (e.g., following wind or ocean currents) to explicitly model vector-based dispersal without altering the underlying architecture.

\paragraph{Spatial Propagation (Regional Context).}
Raw environmental measurements at a single location are often insufficient due to the {Mass Effect} paradigm, where species populations are sustained by the stability of the surrounding meta-community. However, spatial influence is rarely uniform; barriers (e.g., rivers, elevation changes) can decouple geographically close regions.

To model these non-uniform dependencies, we employ a Graph Attention Network (GAT) \citep{velivckovic2017graph} on the grid graph. We define the combined node feature vector at time $\tau$ as $\mathbf{f}_{c,\tau} = [\mathbf{x}_{c,\tau} \parallel \mathbf{y}_{c,\tau}]$, concatenating current covariates with the community state. At each time step $\tau$, we compute the attention coefficients for every edge $(c, v)$:
\begin{equation}
    e_{cv}^\tau = \text{LeakyReLU}\left( \mathbf{a}^\top [\mathbf{W}_{s} \mathbf{f}_{c,\tau} \parallel \mathbf{W}_{s} \mathbf{f}_{v,\tau}] \right),
\end{equation}
where $\mathbf{W}_{s}$ is a learnable projection matrix, $\mathbf{a}$ is the attention vector, and $\parallel$ denotes concatenation. We normalize these coefficients using a softmax function to obtain the final attention weights $\alpha_{cv}^\tau$:
\begin{equation}
    \alpha_{cv}^\tau = \frac{\exp(e_{cv}^\tau)}{\sum_{k \in \mathcal{N}(c) \cup \{c\}} \exp(e_{ck}^\tau)}.
\end{equation}

The regionally contextualized feature vector $\mathbf{s}_{c,\tau}$ is then computed as a weighted sum of the neighbors:
\begin{equation}
    \mathbf{s}_{c,\tau} = \sigma \left( \sum_{v \in \mathcal{N}(c) \cup \{c\}} \alpha_{cv}^\tau \mathbf{W}_{s} \mathbf{f}_{v,\tau} \right).
\end{equation}
To stabilize training, we employ multi-head attention, concatenating the outputs of $K$ independent attention heads. This allows the model to capture different modes of spatial interaction (e.g., one head tracking hydrological connectivity, another tracking thermal gradients).

\paragraph{Temporal Dynamics (Ecological Memory).}
Having established the regional context $\mathbf{s}_{c,\tau}$ at each step, we next model the temporal evolution of these conditions. Ecological drivers often exhibit significant lag effects. We process the sequence of spatial features $\mathbf{S}_{c,t} = [\mathbf{s}_{c, t-W+1}, \dots, \mathbf{s}_{c,t}]$ using a Gated Recurrent Unit (GRU). At each time step $\tau$, the update equations are:
\begin{align}
    \mathbf{z}_\tau &= \sigma(\mathbf{W}_z \mathbf{s}_{c,\tau} + \mathbf{U}_z \mathbf{h}_{c,\tau-1} + \mathbf{b}_z), \\
    \mathbf{r}_\tau &= \sigma(\mathbf{W}_r \mathbf{s}_{c,\tau} + \mathbf{U}_r \mathbf{h}_{c,\tau-1} + \mathbf{b}_r), \\
    \tilde{\mathbf{h}}_\tau &= \tanh(\mathbf{W}_h \mathbf{s}_{c,\tau} + \mathbf{U}_h (\mathbf{r}_\tau \odot \mathbf{h}_{c,\tau-1}) + \mathbf{b}_h), \\
    \mathbf{h}_{c,\tau} &= (1 - \mathbf{z}_\tau) \odot \mathbf{h}_{c,\tau-1} + \mathbf{z}_\tau \odot \tilde{\mathbf{h}}_\tau.
\end{align}
The final hidden state $\mathbf{h}_{c,t}$ represents the \textit{regional ecological memory}. 

\paragraph{Point-Context Fusion.}
To obtain the final embedding for a specific survey location $u$ (located within grid cell $c$), we fuse the high-resolution local environmental covariates $\mathbf{x}_{u,t}$ with the coarse-grained regional context $\mathbf{h}_{c,t-1}$:
\begin{equation}
    \mathbf{h}_{u,t} = \text{MLP}\left( \mathbf{x}_{u,t} \mathbin\Vert \mathbf{h}_{c,t-1} \right).
\end{equation}
The resulting vector $\mathbf{h}_{u,t} \in \mathbb{R}^{d_h}$ serves as the final {spatio-temporal habitat embedding}. This hierarchical approach allows the model to distinguish between critical source habitats and irrelevant neighbors while retaining the local precision required to detect micro-habitat filters.

\subsection{Context-Anchored Latent Alignment}
\label{sec:alignment}

A fundamental challenge in JSDM is the combinatorial explosion of the output space, which scales as $2^S$ for $S$ species. We posit that ecologically valid assemblages are not uniformly distributed but occupy a lower-dimensional manifold constrained by shared functional traits and environmental filters. To capture this intrinsic geometry, we employ a Conditional Variational Autoencoder (CVAE), which leverages the spatio-temporal embedding $\mathbf{h}_{u,t}$ to condition the generation of a compact latent variable $\mathbf{z}$.

\paragraph{Label-Activated Mixture Prior.}
Standard CVAEs assume a unimodal prior $p(\mathbf{z}) = \mathcal{N}(0, I)$. This is ill-suited for ecological data, which is inherently multimodal; different stable communities can exist under distinct conditions. To address this, we adopt the {Label-Activated Mixture Prior} \citep{bai2022cgmvae}. We assign a learnable prototype embedding $\mathbf{w}_s \in \mathbb{R}^{d_z}$ to each species $s$. These prototypes serve as the centroids for species-specific Gaussian components. For a given training sample with observed species set $\mathcal{P}(\mathbf{y}_{u,t}) = \{s : y_{u,t,s} = 1\}$, the prior is defined as a uniform mixture of the components associated with the \textit{present} species:
\begin{equation}
    p_\psi(\mathbf{z} | \mathbf{y}_{u,t}) = \frac{1}{|\mathcal{P}(\mathbf{y}_{u,t})|} \sum_{s \in \mathcal{P}(\mathbf{y}_{u,t})} \mathcal{N}(\mathbf{z} | \boldsymbol{\mu}_s, \mathbf{I} \cdot \sigma^2_{s}).
\end{equation}
This formulation ensures that the prior probability mass is concentrated in regions of the latent space corresponding to valid combinations of the observed species. It creates a ``gravitational field" in latent space, where the structure of the prior changes dynamically based on the observed biodiversity.

\paragraph{Contrastive Alignment via InfoNCE.}
The inference network maps the habitat embedding $\mathbf{h}_{u,t}$ to the parameters of the approximate posterior $q_\phi(\mathbf{z} | \mathbf{h}_{u,t}) = \mathcal{N}(\boldsymbol{\mu}, \text{diag}(\boldsymbol{\sigma}^2))$. To enable gradient descent, we employ the reparameterization trick to sample $\mathbf{z}_{u,t}$.
To ensure that the sampled latent codes cluster meaningfully around their corresponding species modes, we employ a supervised contrastive objective. We first project the sampled latent code to a metric embedding space via a non-linear projection head: $\tilde{\mathbf{z}}_{u,t} = g_{\text{proj}}(\mathbf{z}_{u,t})$.
We treat this projected latent sample $\tilde{\mathbf{z}}_{u,t}$ as the \textit{anchor}. The species prototypes $\mathbf{w}_s$ act as the keys. For a given anchor, prototypes of present species are \textit{positive keys}, while those of absent species are \textit{negative keys}. We minimize the InfoNCE loss:
\begin{equation}
    \mathcal{L}_{\text{con}} = - \sum_{p \in \mathcal{P}(\mathbf{y}_{u,t})} \log \frac{\exp(\text{sim}(\tilde{\mathbf{z}}_{u,t}, \mathbf{w}_p) / \tau)}{\sum_{s=1}^S \exp(\text{sim}(\tilde{\mathbf{z}}_{u,t}, \mathbf{w}_s) / \tau)},
\end{equation}
where $\text{sim}(\cdot, \cdot)$ denotes cosine similarity and $\tau$ is a temperature scalar. By aligning the \textit{stochastic} sample rather than the deterministic encoder output, we ensure that the entire volume of the posterior distribution is mapped to the correct validity region defined by the community prior.

\iffalse
\paragraph{Ecological Interpretability of the Latent Space.}
A key advantage of this formulation is the semantic interpretability of the latent space. Unlike black-box predictors, the learned species prototypes $\mathbf{w}_s$ exist in the same metric space as the environmental embeddings $\mathbf{h}_{u,t}$. The Euclidean distance $||\mathbf{w}_i - \mathbf{w}_j||$ serves as a data-driven proxy for functional similarity; species that cluster together likely share similar fundamental niches or engage in positive biotic interactions (e.g., mutualism). Similarly, the proximity between a prototype $\mathbf{w}_s$ and a habitat embedding $\mathbf{h}_{u,t}$ provides a measure of suitability that explicitly accounts for community context, offering insights beyond simple occupancy probability.
\fi

\subsection{Imbalance-Aware Decoupled Decoding}
\label{sec:decoding}

The third major component of our framework specifically addresses the ``long-tail'' problem inherent in biodiversity data. In standard JSDMs, the decoder typically assumes a symmetric likelihood (e.g., via Binary Cross Entropy), forcing the model to prioritize the majority class (i.e. absences) to minimize global error. Given that most species are rare, this leads to a trivial solution where the model predicts ``absent'' for all rare species, effectively erasing them from the predictions.

To counter this, we introduce an {Imbalance-Aware Decoupled Decoding} strategy. While we utilize a computationally efficient pairwise inner product as the architectural mechanism, we fundamentally alter the \textit{probabilistic definition} of the decoder. By replacing the standard likelihood with an Asymmetric Focusing objective, we introduce a structural modification to the learning dynamics: this acts as a dynamic gradient gating mechanism that \textit{decouples} the optimization of ``easy'' samples (common absences) from ``hard'' samples (rare presences).

\paragraph{Decoder Architecture (Geometric Readout).}
Structurally, we formulate the decoding process as a {pairwise inner product} between the projected latent habitat anchor $\tilde{\mathbf{z}}_{u,t}$ and the species prototypes $\mathbf{w}_s$ (the same prototypes used in the mixture prior). The probability of presence for species $j$ is given by:
\begin{equation}
    \hat{y}_{u,t,j} = \sigma \left( \tilde{\mathbf{z}}_{u,t}^\top \mathbf{w}_j \right),
\end{equation}
where $\sigma(\cdot)$ is the sigmoid function. This design enforces strictly interpretable consistency: the model predicts a species is present if and only if the current spatio-temporal context embeds into that species' specific niche region in the metric space.

\paragraph{Asymmetric Focusing Mechanism.}
To enforce sensitivity to rare species, we integrate the {Asymmetric Loss (ASL)} \citep{ridnik2021asymmetric} directly into the Evidence Lower Bound (ELBO). ASL operates by dynamically down-weighting the contribution of easy negatives. The reconstruction term is defined as:
\begin{equation}
    \mathcal{L}_{\text{ASL}} = \sum_{j=1}^{S} -y_j L_+ - (1-y_j) L_-,
\end{equation}
where the positive and negative probability estimates are decoupled:
\begin{align}
    L_+ &= (1-p)^{\gamma_+} \log(p), \\
    L_- &= (p_m)^{\gamma_-} \log(1-p_m).
\end{align}
Here, $p = \hat{y}_{u,t,j}$ is the predicted probability, and $p_m = \max(p - m, 0)$ is the {shifted probability}.
The hyperparameters $\gamma_+$ and $\gamma_-$ control the focusing strength. By setting $\gamma_- > \gamma_+$ (e.g., $\gamma_-=4, \gamma_+=1$), we aggressively decay the loss contribution of easy negatives (clear absences) once the model is confident ($p < m$). This prevents the vast number of empty survey records from overwhelming the gradient signal derived from rare species occurrences.

\iffalse
\paragraph{Gradient Analysis.}
To understand the efficacy of this decoupling, consider the gradient of the loss with respect to the output logit $z_j$. For a negative sample ($y=0$), the standard Binary Cross Entropy (BCE) gradient is simply $\frac{\partial \mathcal{L}_{\text{BCE}}}{\partial z} = p$. Even if the model predicts a low probability (e.g., $p=0.01$), the accumulated gradient from thousands of negative samples can be substantial, drowning out the sparse signal from positives.
In contrast, the gradient for ASL includes a dampening factor:
\begin{equation}
    \frac{\partial \mathcal{L}_{\text{ASL-}}}{\partial z} \approx (p_m)^{\gamma_-} \cdot p.
\end{equation}
If $p < m$, the shifted probability $p_m = 0$, and the gradient becomes exactly zero. This creates a ``hard margin'' effect, effectively silencing the noise from abundant easy negatives and redirecting model capacity toward the difficult tail species.
\fi

\paragraph{Optimization.}
Our model is trained end-to-end by minimizing a unified objective that balances generative fidelity, latent structure alignment, and robust long-tail handling. 

%\paragraph{Unified Objective.}
The total objective $\mathcal{L}_{\text{total}}$ combines the Imbalance-Aware ELBO and the auxiliary Contrastive Alignment loss.
\begin{equation}
    \mathcal{L}_{\text{ELBO}} = \mathbb{E}_{q}[\mathcal{L}_{\text{ASL}}(\mathbf{y}, \hat{\mathbf{y}})] - \beta \text{KL}\left( q_\phi(\mathbf{z} | \mathbf{h}_{u,t}) \parallel p_\psi(\mathbf{z} | \mathbf{y}_{u,t}) \right),
\end{equation}
where $\beta$ weights the KL divergence term. The KL term forces the variational posterior (conditioned on the environment) to align with the {community-derived mixture prior} (conditioned on species labels). We approximate this divergence using the log-sum-exp upper bound to efficiently handle the Gaussian Mixture components defined by the species prototypes. The final loss function is:
\begin{equation}
    \mathcal{L}_{\text{total}} = -\mathcal{L}_{\text{ELBO}} + \alpha \mathcal{L}_{\text{con}}.
\end{equation}

\begin{figure}[t]
    \centering
    \includegraphics[width=1.0\linewidth]{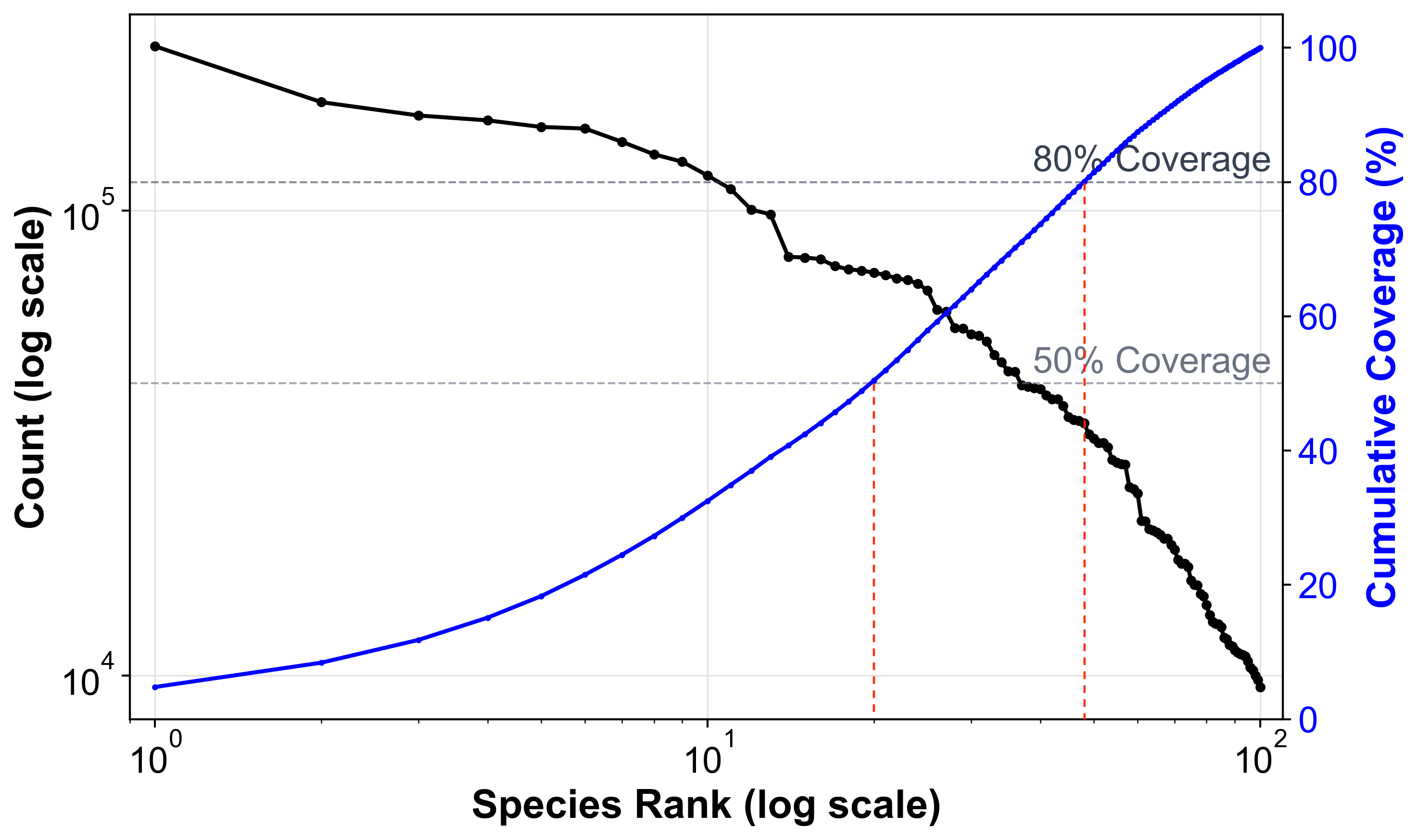} 
    \caption{\textbf{The Long-Tail of Biodiversity.} Rank-frequency distribution of the top 100 species in the eBird dataset (log-log scale). The black line denotes species counts, while the blue curve shows cumulative coverage. The vertical dashed lines highlight extreme class imbalance: the top 20 species alone contribute 50\% of the data, while the top 48 species account for 80\%, leaving a long tail of rare specialists.}
    \label{fig:long_tail}
\end{figure}

\begin{table*}[th]
\centering
\caption{Performance comparison on the eBird dataset. STELLAR consistently outperforms baselines in F1 scores, Precision@1, and Recall. While C-GMVAE achieves slightly higher PR-AUC through a conservative prediction strategy, STELLAR demonstrates superior capability in detecting rare species (Macro Recall), leading to the best overall balance between precision and coverage. \textit{Bold indicates best performance.}}
\label{tab:main_results}
\begin{tabular}{l|ccc|ccc}
\toprule
\textbf{Method} & \textbf{Example-F1} & \textbf{Micro-F1} & \textbf{Macro-F1} & \textbf{Precision@1} & \textbf{Macro Recall} & \textbf{Macro PR-AUC} \\ \midrule
DMSE        & 0.3302 & 0.4423 & 0.2707 & 0.4050 & 0.2575 & 0.1860 \\
MulSupCon   & 0.2898 & 0.4148 & 0.2348 & 0.6594 & 0.1712 & 0.4207 \\
C-GMVAE     & 0.3993 & 0.5210 & 0.3979 & 0.6943 & 0.3123 & \textbf{0.4901} \\ 
SINR        & 0.3770 & 0.4060 & 0.2649 & 0.5671 & 0.4077 & 0.2233 \\ 
LabelKAN    & 0.4148 & 0.5322 & 0.4208 & 0.6932 & 0.3420 & 0.4890 \\ 
\midrule
STELLAR  & \textbf{0.4895} & \textbf{0.5701} & \textbf{0.4766} & \textbf{0.7019} & \textbf{0.4911} & 0.4816 \\ \bottomrule
\end{tabular}
\end{table*}

\begin{figure*}[htbp]
  \centering
  \begin{subfigure}[b]{0.55\textwidth}
    \includegraphics[width=\linewidth]{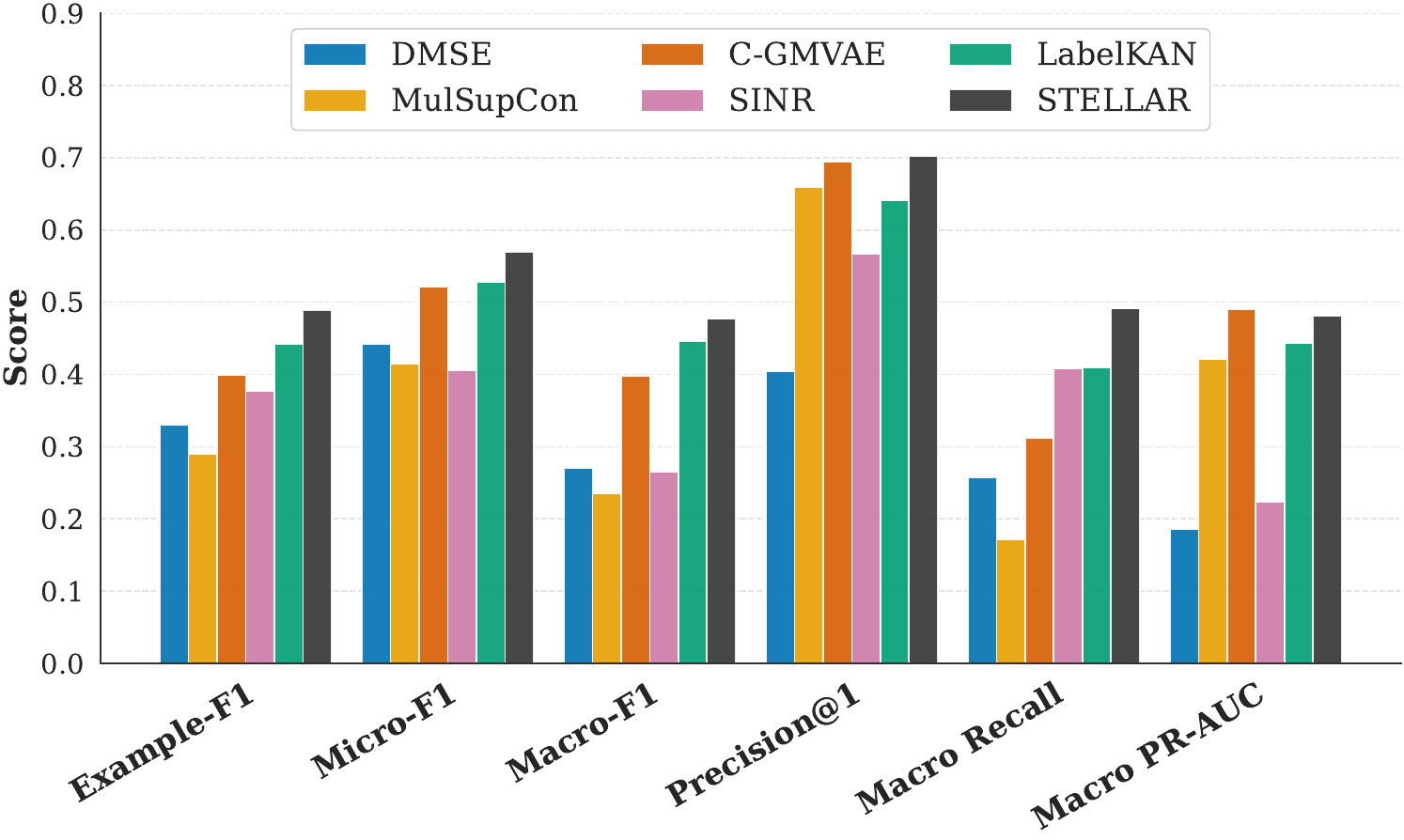}
    %\caption{Bar chart comparison of model performance across multiple metrics.}
    %\label{fig:bar_chart}
  \end{subfigure}
  \hfill
  \begin{subfigure}[b]{0.44\textwidth}
    \includegraphics[width=\linewidth]{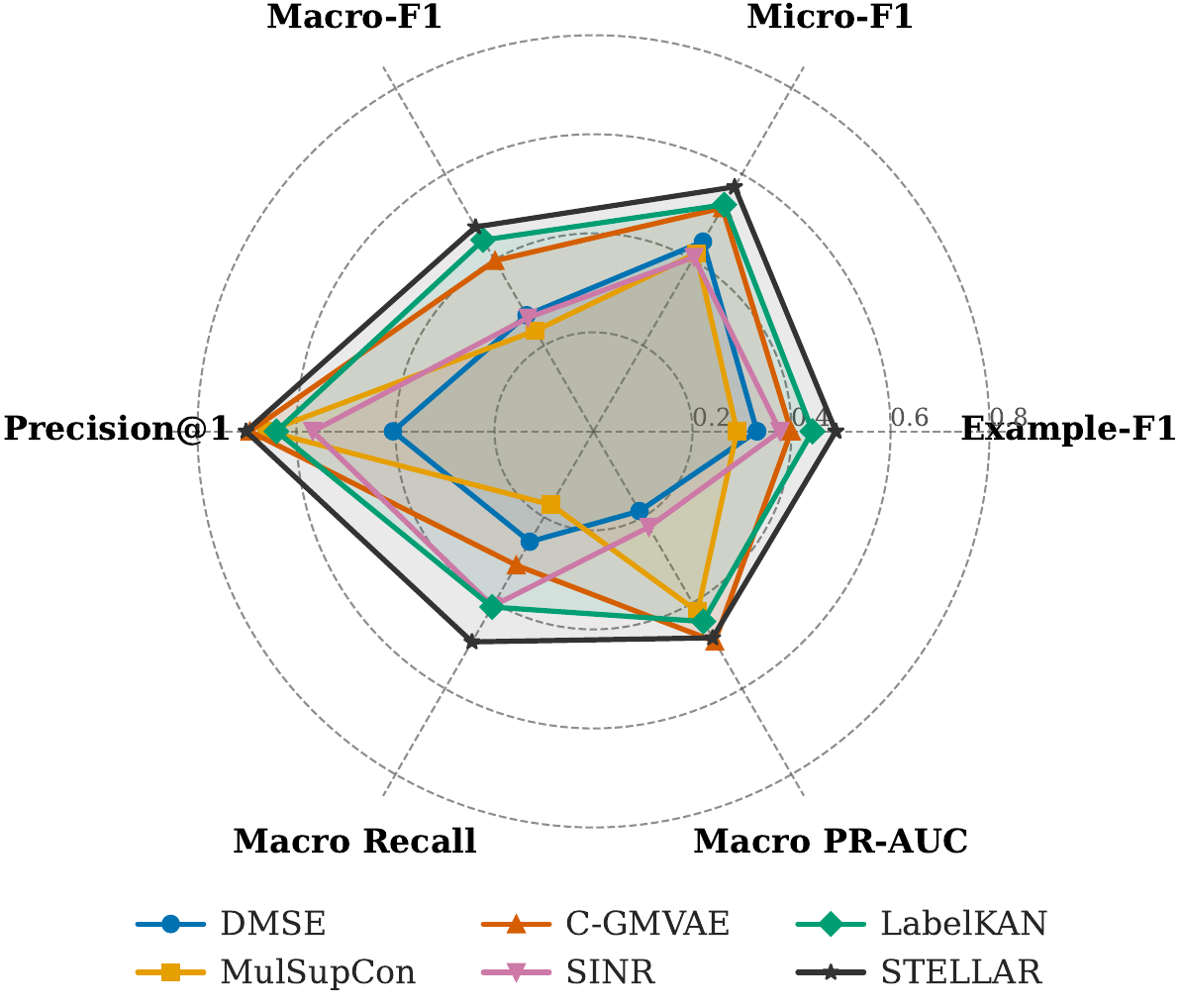}
    %\caption{Radar chart comparison of model performance across the same metrics.}
    %\label{fig:radar_chart}
  \end{subfigure}
  %\vspace{-5mm}
  \caption{Visual comparison of model capabilities. The plots illustrate STELLAR's superior handling of long-tailed distributions. By effectively balancing Precision and Recall, our framework mitigates the mode collapse observed in baselines, ensuring robust detection of both common and rare species.}
  \label{fig:model_comparison}
\end{figure*}

\section{Experiments}
\label{sec:experiments}

%We evaluate our framework on the \textbf{eBird} dataset, the world's largest citizen-science biodiversity monitoring project. %Our experimental design aims to answer two key questions: (1) Does the framework effectively mitigate the ``long-tail'' bias inherent in community science data? (2) Do the structural components (Spatio-Temporal Context and Geometric Alignment) contribute meaningfully to ecological generalization?

\subsection{Dataset Curation}
%\subsubsection{The eBird Dataset.} 
We evaluate our framework on the \textbf{eBird} dataset, the world's largest citizen-science biodiversity monitoring project. eBird is a crowd-sourced bird observation dataset \citep{sullivan2009ebird}. We curate a subset of the eBird Basic Dataset (EBD) covering North America, restricting our analysis to the top $S=100$ most frequently observed species. The final dataset comprises {434,352 checklists}. 
Each checklist is associated with a {104-dimensional feature vector} describing the sampling context. %This vector concatenates environmental and observer covariates: {location} (2 dims), {observer metadata} (6 dims: e.g., effort hours, time), {climate} (30 dims), {topology} (6 dims), and {land cover} (60 dims) derived from the National Land Cover Dataset (NLCD) \citep{homer2015completion}.
Consistent with ecological monitoring, the data exhibits a severe power-law distribution (see Figure \ref{fig:long_tail}). The top {20 dominant generalists} account for {50\%} of all detections. Conversely, the ``tail'' comprises 80 specialist species, with the rarest appearing in roughly {2\%} of checklists.

\subsection{Experimental Settings}
\paragraph{Data Splitting.}
To rigorously evaluate the modeling of regional connectivity and historical dynamics, we employ a \textit{temporal splitting strategy}. We utilize data from 2014--2018 (319,078 checklists) as \textit{historical context} to construct the regional spatio-temporal priors. The subsequent data from 2019 (115,274 checklists) serves as the target set for model development and is randomly partitioned into training (80\%), validation (10\%), and testing (10\%) subsets.

\paragraph{Baselines.}
We benchmark STELLAR against five state-of-the-art methods representing diverse JSDM categories. In the embedding-based domain, \textbf{DMSE} \citep{ijcai2017p509} maps environmental features and species into a shared latent space, modeling correlations via a multivariate probit framework. \textbf{MulSupCon} \citep{zhang2024multi} extends discriminative representation learning to multi-label tasks by weighting contrastive positive samples based on label overlap. Representing generative approaches, \textbf{C-GMVAE} \citep{bai2022cgmvae} conditions a VAE on environmental features using a Gaussian Mixture prior to capture multimodal distributions. For spatial representation, \textbf{SINR} \citep{cole2023spatial} utilizes implicit neural representations to learn continuous coordinate-based functions for global interpolation. Finally, \textbf{LabelKAN} \citep{grimson2026labelkan} serves as a structure learning baseline, employing Kolmogorov-Arnold Networks to explicitly model high-order label correlations through learnable activation functions.

\begin{figure*}[th]
    \centering
    \includegraphics[width=1.0\linewidth]{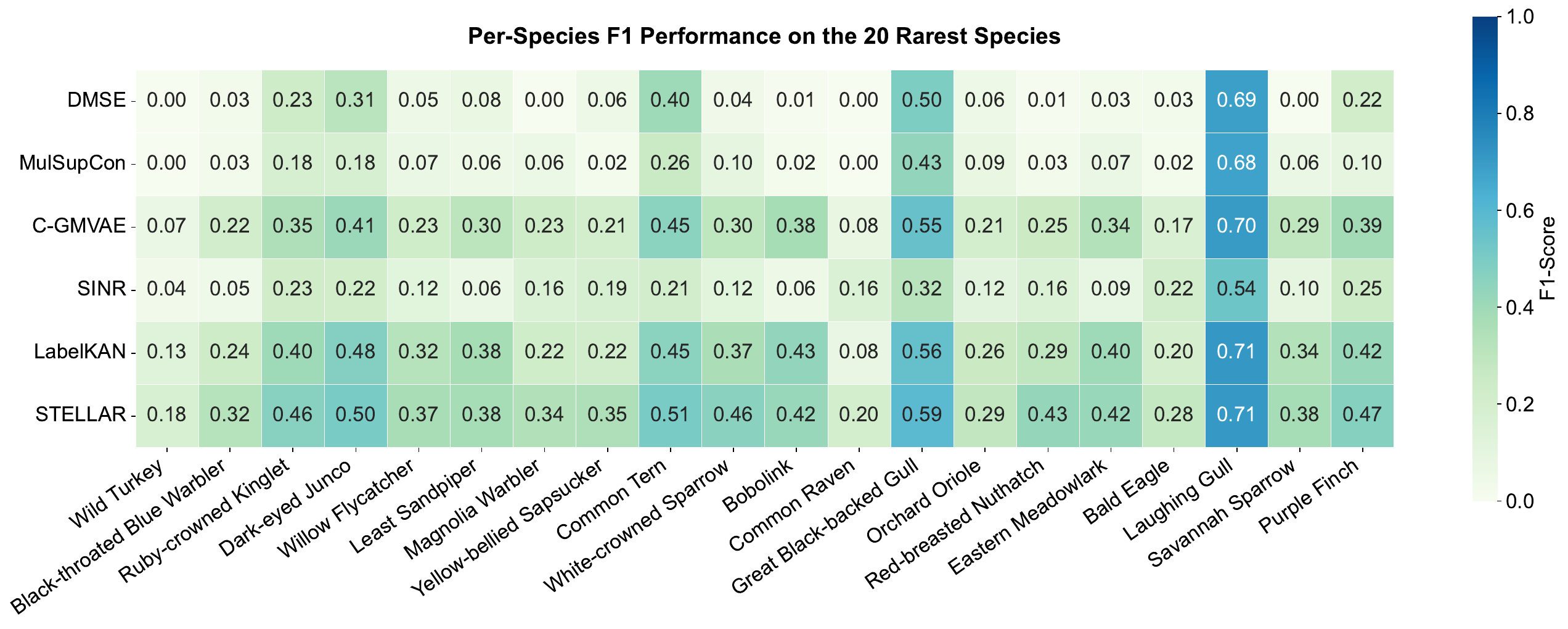} 
    \caption{Per-species F1 performance heatmap on the 20 rarest species (sorted by total support). The x-axis represents bird species ordered by increasing rarity (lowest support on the left). Darker cells indicate higher F1 scores. While baselines like DMSE and MulSupCon exhibit severe mode collapse (predicting zero for most rare species), STELLAR maintains robust detection capabilities across the long tail, successfully recovering species that are effectively invisible to other models.}
    \label{fig:f1_heatmap}
\end{figure*}

\paragraph{Metrics.}
We evaluate model performance using comprehensive multi-label classification metrics. We report \textbf{Micro-F1}, \textbf{Macro-F1}, and \textbf{Example-F1} to assess the balance between precision and recall at different aggregation levels. Crucially, \textbf{Macro-F1} and \textbf{Macro Recall} serve as the primary indicators for our long-tailed setting; by averaging performance per species regardless of support, they explicitly penalize models that fail to detect rare species. Additionally, we employ \textbf{Macro PR-AUC} and \textbf{Precision@1} to evaluate the quality of the predictive ranking and the accuracy of the top-confidence predictions, respectively.

\paragraph{Implementation.} We implemented our model STELLAR using PyTorch and the PyG (PyTorch Geometric) library \citep{fey2025pyg20scalablelearning}. All experiments were conducted on a workstation equipped with 4 NVIDIA GeForce RTX 3090 GPUs. Due to space constraints, we defer detailed specifications regarding data preprocessing, model architecture, and training strategy to the Appendix~\ref{appx:technical}.

\subsection{Results and Analysis}

Table \ref{tab:main_results} and Figure \ref{fig:model_comparison} present the comparative performance on the eBird dataset. 
\textbf{STELLAR} achieves state-of-the-art results across the majority of metrics, surpassing the strongest baseline (LabelKAN) by a significant margin in \textbf{Macro-F1 (0.477 vs. 0.421)} and \textbf{Macro Recall (0.491 vs. 0.342)}.
Notably, our model attains the highest \textbf{Precision@1 (0.702)}, demonstrating that our imbalance-aware decoding mechanism effectively filters false positives even while maximizing recall.
We observe that \textbf{C-GMVAE} yields a marginally higher \textbf{Macro PR-AUC (0.490)} compared to STELLAR (0.482). This slight difference arises because PR-AUC favors conservative models that assign high confidence to a narrow set of ``easy" head classes. C-GMVAE achieves this ranking stability by sacrificing coverage, as evidenced by its significantly lower \textbf{Macro Recall (0.312)}. 
In contrast, STELLAR extends its predictive capacity to the difficult ``long tail" of rare species. While ranking these uncertain rare events causes a minor drop in PR-AUC, it results in a massive \textbf{+57\% relative improvement in Macro Recall} over C-GMVAE, proving that STELLAR offers a far more comprehensive solution for biodiversity monitoring.

\paragraph{Rare Species Detection Analysis.}
To further investigate the model's capability in the extreme long-tail regime, we visualize the per-species F1 scores for the 20 rarest species in Figure \ref{fig:f1_heatmap}.
As illustrated, traditional embedding-based methods such as \textbf{DMSE} and \textbf{MulSupCon} suffer from severe mode collapse, failing to detect the majority of these species (indicated by the prevalence of $0.00$ scores). This confirms that without explicit handling of class imbalance, discriminative models tend to bias heavily toward majority classes to minimize aggregate loss.
While the generative baseline \textbf{C-GMVAE} improves detection rates, it still exhibits inconsistency, missing several species entirely.
In contrast, \textbf{STELLAR} demonstrates superior robustness, providing consistent non-zero F1 scores across the entire set of rare species. This performance gain is directly attributable to our \textit{Imbalance-Aware Decoupled Decoding} strategy, which prevents the gradients from dominant head classes from suppressing the learning of sparse tail signals. These results highlight STELLAR's potential for conservation applications, where detecting rare and threatened species is often the primary objective.

We also give qualitative visualizations of the spatial predictions for the 20 rarest species in the Appendix \ref{app:spatial_vis}. These distribution maps reveal that STELLAR captures rare events more effectively than baselines and more accurately reconstructs fine-grained spatial structures.

%\subsection{Ablation Study}
\paragraph{Ablation Study.}
We conduct ablation study to quantify the contribution of each core component in STELLAR by analyzing the relative performance decline upon its removal.
First, removing the Imbalance-Aware Decoupled Decoding results in a sharp 26.2\% decrease in Macro Recall and a 10.3\% drop in Macro-F1, verifying its critical role in recovering rare species.
Next, replacing the Spatio-Temporal Encoder with a standard MLP causes a severe degradation, with a 30.4\% drop in Macro Recall and an 11.7\% decline in Macro-F1, confirming that explicit spatial modeling is important.
Finally, simplifying the Context-Anchored Latent Alignment to a univariate Gaussian prior leads to a 5.6\% decline in Macro Recall and a 3.1\% drop in Macro-F1, highlighting the necessity of multimodal priors for complex distributions.

%We conduct ablation study to identify the contribution of major compnents: (1) We removing the spatio-temporal encoder with a simple MLP like C-GMVAE, the Example-F1, Micro-F1, Macro-F1, Precision@1, Macro Recall, and Macro PR-AUC drop to 0.4148, 0.5322, 0.4208, 0.6932, 0.3420, and 0.4890 respectively; (2) We replace the multi-Gaussian wit strict univariance Gaussian in the context-anchored latent alignment, the Example-F1, Micro-F1, Macro-F1, Precision@1, Macro Recall, and Macro PR-AUC drop to 0.4678, 0.5310, 0.4617, 0.6798, 0.4638, and 0.4709 respectively; and (3) We remove the Imbalance-Aware Decoupled Decoding module, the Example-F1, Micro-F1, Macro-F1, Precision@1, Macro Recall, and Macro PR-AUC drop to 0.4293, 0.5287, 0.4273, 0.6547, 0.3623, and 0.4596 respectively.
%We further evaluate ablated variants to isolate component gains. Replacing the Imbalance-Aware Decoding with standard Binary Cross Entropy causes the most severe drop in \textbf{Macro-F1 (-18.2\%)}, confirming that standard losses ignore the tail. Similarly, removing the spatio-temporal encoder degrades Micro-F1 (-5.4\%), validating the importance of regional context for generalist species.

\section{Conclusion}
In this paper, we proposed STELLAR, a novel framework designed to address the challenges of large-scale biodiversity monitoring. Our approach integrates three core components: a GNN-RNN Encoder for capturing context-rich biophysical embeddings; a Context-Anchored Latent Alignment mechanism for multimodal structure learning; and an Imbalance-Aware Decoupled Decoding module for robust long-tail prediction. Extensive experiments and analysis confirm the superior performance of our model against state-of-the-art baselines, establishing STELLAR as a comprehensive and effective solution for fine-grained species distribution modeling.

%\section*{Contributions}

\section*{Acknowledgements}
This work was supported by the National Natural Science Foundation of China (Grant No. 62506090) and the National Key R\&D Program of China (Grant No. 2025YFF0523900).

% \newpage
%% The file named.bst is a bibliography style file for BibTeX 0.99c
\bibliographystyle{named}
\bibliography{ijcai26}

@article{mcinnes2018umap,
  title={Umap: Uniform manifold approximation and projection for dimension reduction},
  author={McInnes, Leland and Healy, John and Melville, James},
  journal={arXiv preprint arXiv:1802.03426},
  year={2018}
}

@misc{fey2025pyg20scalablelearning,
      title={PyG 2.0: Scalable Learning on Real World Graphs}, 
      author={Matthias Fey and Jinu Sunil and Akihiro Nitta and Rishi Puri and Manan Shah and Blaž Stojanovič and Ramona Bendias and Alexandria Barghi and Vid Kocijan and Zecheng Zhang and Xinwei He and Jan Eric Lenssen and Jure Leskovec},
      year={2025},
      eprint={2507.16991},
      archivePrefix={arXiv},
      primaryClass={cs.LG},
      url={https://arxiv.org/abs/2507.16991}, 
}

@inproceedings{ijcai2017p509,
  author    = {Di Chen and Yexiang Xue and Daniel Fink and Shuo Chen and Carla P. Gomes},
  title     = {Deep Multi-species Embedding},
  booktitle = {Proceedings of the Twenty-Sixth International Joint Conference on Artificial Intelligence, {IJCAI-17}},
  pages     = {3639--3646},
  year      = {2017},
  doi       = {10.24963/ijcai.2017/509},
  url       = {https://doi.org/10.24963/ijcai.2017/509},
}

@inproceedings{zhang2024multi,
  title={Multi-label supervised contrastive learning},
  author={Zhang, Pingyue and Wu, Mengyue},
  booktitle={Proceedings of the AAAI conference on artificial intelligence},
  volume={38},
  number={15},
  pages={16786--16793},
  year={2024}
}

@article{homer2015completion,
  title={Completion of the 2011 National Land Cover Database for the conterminous United States--representing a decade of land cover change information},
  author={Homer, Collin and Dewitz, Jon and Yang, Limin and Jin, Suming and Danielson, Patrick and Xian, George and Coulston, John and Herold, Nathaniel and Wickham, James and Megown, Kevin},
  journal={Photogrammetric Engineering \& Remote Sensing},
  volume={81},
  number={5},
  pages={345--354},
  year={2015},
  publisher={Elsevier}
}

@article{sullivan2009ebird,
  title={eBird: A citizen-based bird observation network in the biological sciences},
  author={Sullivan, Brian L and Wood, Christopher L and Iliff, Marshall J and Bonney, Rick E and Fink, Daniel and Kelling, Steve},
  journal={Biological conservation},
  volume={142},
  number={10},
  pages={2282--2292},
  year={2009},
  publisher={Elsevier}
}

@inproceedings{lin2017focal,
  title={Focal loss for dense object detection},
  author={Lin, Tsung-Yi and Goyal, Priya and Girshick, Ross and He, Kaiming and Doll{\'a}r, Piotr},
  booktitle={Proceedings of the IEEE international conference on computer vision},
  pages={2980--2988},
  year={2017}
}

@inproceedings{menon2020long,
  title={Long-tail learning via logit adjustment},
  author={Menon, Aditya Krishna and Jayasumana, Sadeep and Rawat, Ankit Singh and Jain, Himanshu and Veit, Andreas and Kumar, Sanjiv},
  booktitle={International Conference on Learning Representations},
  year={2021}
}

@article{golding2016fast,
  title={Fast and flexible Bayesian species distribution modelling using Gaussian processes},
  author={Golding, Nick and Purse, Bethan V},
  journal={Methods in Ecology and Evolution},
  volume={7},
  number={5},
  pages={598--608},
  year={2016},
  publisher={Wiley Online Library}
}

@inproceedings{ridnik2021asymmetric,
  title={Asymmetric loss for multi-label classification},
  author={Ridnik, Tal and Ben-Baruch, Emanuel and Zamir, Nadav and Noy, Asaf and Friedman, Itamar and Protter, Matan and Zelnik-Manor, Lihi},
  booktitle={Proceedings of the IEEE/CVF international conference on computer vision},
  pages={82--91},
  year={2021}
}

@article{ovaskainen2011making,
  title={Making more out of sparse data: hierarchical modeling of species communities},
  author={Ovaskainen, Otso and Soininen, Janne},
  journal={Ecology},
  volume={92},
  number={2},
  pages={289--295},
  year={2011},
  publisher={Wiley Online Library}
}

@article{velivckovic2017graph,
  title={Graph attention networks},
  author={Veli{\v{c}}kovi{\'c}, Petar and Cucurull, Guillem and Casanova, Arantxa and Romero, Adriana and Lio, Pietro and Bengio, Yoshua},
  journal={arXiv preprint arXiv:1710.10903},
  year={2017}
}

@article{gomes2019computational,
  title={Computational sustainability: Computing for a better world and a sustainable future},
  author={Gomes, Carla and Dietterich, Thomas and others},
  journal={Communications of the ACM},
  volume={62},
  number={9},
  pages={56--65},
  year={2019}
}

@article{elith2009species,
  title={Species distribution models: ecological explanation and prediction across space and time},
  author={Elith, Jane and Leathwick, John R},
  journal={Annual review of ecology, evolution, and systematics},
  volume={40},
  pages={677--697},
  year={2009}
}

@article{wisz2013role,
  title={The role of biotic interactions in shaping distributions and realised assemblages of species: implications for species distribution modelling},
  author={Wisz, Mary S and Pottier, Julien and Kissling, W Daniel and others},
  journal={Biological reviews},
  volume={88},
  number={1},
  pages={15--30},
  year={2013}
}

@inproceedings{chen2018end,
  title={End-to-End Learning for the Deep Multivariate Probit Model},
  author={Chen, Di and Xue, Yexiang and Gomes, Carla},
  booktitle={International Conference on Machine Learning (ICML)},
  pages={932--941},
  year={2018},
  organization={PMLR}
}

@inproceedings{bai2020disentangled,
  title={Disentangled Variational Autoencoder based Multi-Label Classification with Covariance-Aware Multivariate Probit Model},
  author={Bai, Junwen and Kong, Shufeng and Gomes, Carla},
  booktitle={International Joint Conference on Artificial Intelligence (IJCAI)},
  year={2020}
}

@inproceedings{fan2022gnn,
  title={A GNN-RNN Approach for Harnessing Geospatial and Temporal Information: Application to Crop Yield Prediction},
  author={Fan, Joshua and Bai, Junwen and Li, Zhiyun and Ortiz-Bobea, Ariel and Gomes, Carla P},
  booktitle={Proceedings of the AAAI Conference on Artificial Intelligence},
  volume={36},
  number={11},
  pages={11873--11881},
  year={2022}
}

@inproceedings{cole2023spatial,
  title={Spatial Implicit Neural Representations for Global-Scale Species Mapping},
  author={Cole, Elijah and Van Horn, Grant and Lange, Christian and others},
  booktitle={International Conference on Machine Learning (ICML)},
  pages={6078--6095},
  year={2023},
  organization={PMLR}
}

@article{ovaskainen2017make,
  title={How to make more out of sparse data: hierarchical modeling of species communities},
  author={Ovaskainen, Otso and Tikhonov, Gleb and others},
  journal={Ecology},
  volume={98},
  number={10},
  pages={2626--2643},
  year={2017}
}

@inproceedings{grimson2026labelkan,
  title={LabelKAN - Kolmogorov-Arnold Networks for Inter-Label Learning: Avian Community Learning},
  author={Grimson, Marc and Fan, Joshua and Davis, Courtney and van Bramer, Dylan and Fink, Daniel and Gomes, Carla},
  booktitle={Proceedings of the AAAI Conference on Artificial Intelligence},
  year={2026}
}

@inproceedings{bai2022cgmvae,
  title     = {Gaussian Mixture Variational Autoencoder with Contrastive Learning for Multi-Label Classification},
  author    = {Bai, Junwen and Kong, Shufeng and Gomes, Carla P.},
  booktitle = {Proceedings of the 39th International Conference on Machine Learning (ICML)},
  series    = {Proceedings of Machine Learning Research},
  volume    = {162},
  year      = {2022}
}

\clearpage
\appendix

\section{Additional Result Analysis}

\paragraph{Latent Structure Analysis.} To investigate the semantic structure learned by STELLAR, we visualize the embeddings of 100 representative species using Uniform Manifold Approximation and Projection (UMAP) \citep{mcinnes2018umap}. We apply clustering to these projected embeddings to identify latent groupings. As detailed in Appendix \ref{app:embeddings}, the visualization reveals that STELLAR learns a highly structured latent space where species with similar ecological niches and co-occurrence patterns form distinct, compact clusters.

\paragraph{ASL Hyperparameter Sensitivity.}
We further analyze the sensitivity of STELLAR to the asymmetric loss (ASL) hyperparameters. In particular, we vary $\gamma_{-} \in \{2,4\}$ while keeping $\gamma_{+}$ fixed, and observe a clear and interpretable monotonic trade-off. As $\gamma_{-}$ increases, the loss places progressively less emphasis on easy negative examples, which is consistent with the original motivation of ASL in highly imbalanced multi-label settings. This shift improves Macro Recall from $0.491$ to $0.537$, indicating stronger sensitivity to positive labels, especially for harder or rarer classes. At the same time, Macro-F1 decreases from $0.477$ to $0.447$, reflecting the expected precision--recall trade-off induced by more aggressive suppression of easy negatives. In practice, it allows practitioners to tune the operating point of the model according to downstream conservation priorities.
For example, applications that prioritize missing as few candidate species as possible may prefer a larger $\gamma_{-}$, whereas applications requiring a more balanced precision--recall profile may prefer a smaller value.

\paragraph{Robustness under Temporal Shifts.}
To examine robustness to temporal context length, we ablate the size of the historical context window used by the model. We find that using $W=3\,\mathrm{yr}$ yields the same Macro Recall as using $W=5\,\mathrm{yr}$ ($0.491$ vs.\ $0.491$). Even when the context is reduced to only $W=1\,\mathrm{yr}$, performance drops by only $3.3\%$, from $0.491$ to $0.475$. This relatively small degradation indicates that STELLAR remains effective even when only minimal historical information is available. More broadly, the result suggests that the model does not rely on a narrowly tuned temporal horizon and is resilient to moderate temporal perturbations in the input context. Such behavior is desirable for ecological forecasting, where data availability may vary across regions and years, and where gradual climate-driven shifts can alter the usefulness of long historical windows. Overall, the ablation supports the view that STELLAR captures temporally stable ecological structure rather than depending excessively on a specific choice of context length.

\paragraph{Computational Cost.}
We also evaluate computational efficiency by measuring the training time per epoch on RTX 3090 GPU. Under this setting, STELLAR requires only $49\,\mathrm{s}$ per epoch, making it substantially faster than DMSE ($509\,\mathrm{s}$), MulSupCon ($117\,\mathrm{s}$), C-GMVAE ($72\,\mathrm{s}$), and LabelKAN ($1494\,\mathrm{s}$). Relative to LabelKAN, which is our strongest baseline in terms of predictive performance, STELLAR is approximately $30\times$ faster. This efficiency is important in practice because it directly affects the feasibility of large-scale hyperparameter tuning, repeated ablation studies, and retraining under updated data snapshots. The grid backbone (Appendix~\ref{appx:technical}) is shared across all checklists per time step, enabling efficient scaling. Among the compared methods, only SINR is faster ($0.75\,\mathrm{s}$ per epoch), but this speed comes with a clear predictive trade-off, achieving lower Macro Recall ($0.408$ vs.\ $0.491$).
These results show that STELLAR offers a favorable balance between accuracy and efficiency, rather than optimizing one at the expense of the other.

\section{Deployment and Collaboration.}
Beyond predictive performance, STELLAR is designed to support practical conservation workflows. When the model detects declining occurrence relative to the historical baseline, it can directly flag the corresponding spatial zone for follow-up surveys and habitat protection. This is aligned with the broader role of species distribution modeling (SDM) in conservation planning, where model outputs are used to prioritize limited monitoring and intervention resources. As one concrete example of this workflow, SDM-based analyses have identified that more than $25\%$ of priority grassland-species habitat lies outside protected areas, motivating new conservation action. STELLAR fits naturally into this decision pipeline by providing spatially resolved, temporally informed predictions that can help identify emerging areas of concern.

Our collaboration with the Cornell Lab of Ornithology is central to this design. This collaboration encompassed three main aspects:
(1) identifying rare-species prediction as a key conservation bottleneck;
(2) curating the $104$-dimensional covariate set so that it is ecologically grounded rather than purely convenience-driven; and
(3) validating model behavior against known species ranges.
These interactions helped ensure that the modeling choices were informed by domain relevance, not only by benchmark performance. In addition, the UMAP visualization in Figure \ref{fig:umap_vis} shows that the learned latent structure recovers coherent ecological guilds, including woodland species, migratory warblers, and aquatic birds. The fact that these groupings are ecologically interpretable provides further qualitative evidence that the model is capturing meaningful biological structure rather than only dataset-specific correlations.

\section{Cross-Domain Applicability.}
Although STELLAR is evaluated here on bird occurrence data, its formulation does not rely on bird-specific assumptions. At the architectural level, the GNN operates on a spatial graph whose nodes are represented by tabular covariates, a design that is not tied to any single taxonomic group. Likewise, the mixture-prior formulation is a general mechanism for modeling structured latent variation in multi-label communities, and does not depend on specific label semantics. As a result, the framework can in principle be transferred to other biodiversity monitoring settings that involve spatially structured covariates and multi-species prediction.

For example, applying STELLAR to plants using iNaturalist records or to mammals using GBIF records would primarily require replacing the task-specific observation data and adapting the covariate configuration accordingly. Importantly, the current feature space already includes $104$ variables describing climate, land cover, and elevation, which are broadly available at global scale for terrestrial ecosystems.
This makes the framework naturally compatible with other terrestrial taxa, provided that suitable occurrence labels and graph construction procedures are available. We therefore view STELLAR not as a bird-only model, but as a general approach to large-scale, multi-label species community modeling under spatial and environmental structure.

\section{Technical Details of STELLAR}
\label{appx:technical}

\subsection{Implementation Details}
We implemented STELLAR using PyTorch and the PyG (PyTorch Geometric) library \citep{fey2025pyg20scalablelearning}. All experiments were conducted on a workstation equipped with 4 NVIDIA GeForce RTX 3090 GPUs.

\paragraph{Dataset Curation.}
%\subsubsection{The eBird Dataset.} 
We evaluate our framework on the \textbf{eBird} dataset, the world's largest citizen-science biodiversity monitoring project. eBird is a crowd-sourced bird observation dataset \citep{sullivan2009ebird}. We curate a subset of the eBird Basic Dataset (EBD) covering North America, restricting our analysis to the top $S=100$ most frequently observed species. The final dataset comprises {434,352 checklists}. 
Each checklist is associated with a {104-dimensional feature vector} describing the sampling context. This vector concatenates environmental and observer covariates: {location} (2 dims), {observer metadata} (6 dims: e.g., effort hours, time), {climate} (30 dims), {topology} (6 dims), and {land cover} (60 dims) derived from the National Land Cover Dataset (NLCD) \citep{homer2015completion}.

\paragraph{Data Preprocessing.}
To capture spatial dependencies, we discretized the study area into grid cells of size $0.05^{\circ} \times 0.05^{\circ}$ and constructed a spatial graph using a King's graph topology (8-neighbor connectivity). A coarser $0.1^\circ \times 0.1^\circ$ grid yielded consistently lower performance (e.g., Macro Recall $0.477$ vs.\ $0.491$), which confirmed that that our choice captured meaningful local habitat variation without introducing excessive sparsity. Continuous environmental covariates were Z-score normalized using statistics derived strictly from the training years (2014--2018) to prevent data leakage. Spatial coordinates were transformed into 3D Cartesian coordinates $(x, y, z)$ to handle spherical distortions, while temporal information was encoded using cyclical sine and cosine functions. Missing historical observations were imputed via a spatial smoothing strategy, which prioritizes interpolation from concurrent spatial neighbors before falling back to global averages.
%We adopted a strict temporal splitting strategy: data from 2014 to 2018 served as the historical context for the encoder (without loss calculation), while the 2019 data were split into training ($80\%$), validation ($10\%$), and testing ($10\%$) sets for the target prediction task.

\paragraph{Model Architecture.}
The \textit{Spatio-Temporal Encoder} consists of a 2-layer GATv2 architecture with residual connections and LayerNorm. We employed $k=2$ attention heads with a hidden dimension of $256$. Temporal evolution is modeled by a single-layer GRU ($L_{gru}=1$), producing a final spatio-temporal context embedding of dimension $d_{ctx} = 128$.
The \textit{Context-Anchored Latent Alignment} (C-GMVAE) operates with a latent dimension of $d_z = 64$. Species labels are projected into an embedding space of size $2048$.

\paragraph{Training Strategy.}
The model was trained end-to-end using the Adam optimizer with an initial learning rate of $1 \times 10^{-4}$. We utilized a Cosine Annealing scheduler configured with $T_{0}=200$ and $T_{mult}=2$, decaying the learning rate to a minimum of $1 \times 10^{-7}$. The maximum training duration was set to $200$ epochs with a batch size of $256$.
To prevent overfitting, we applied Dropout with $p=0.5$ to the fully connected layers of the C-GMVAE, while the GNN-RNN module was trained without dropout ($p=0.0$).
The total loss $\mathcal{L}_{total}$ was computed as a weighted sum of the reconstruction loss ($\lambda_{rec}=0.5$), KL-divergence ($\lambda_{kl}=6.0$), and supervised contrastive loss ($\lambda_{con}=1.0$), with the contrastive temperature parameter set to $\tau=1.0$.

\paragraph{Prior Construction at Inference Time.}
At test time, no prior construction is required: STELLAR uses the posterior mean $\mu$ directly as a deterministic latent code ($\mathbf{z} = \mathbf{\mu}$, no sampling). The inference network $q_\phi(\mathbf{z} | \mathbf{h}_{u,t})$ maps the habitat embedding to $\mu$ independently of species labels, so predictions depend only on the observed environmental context $h_{u,t}$. The label-conditioned mixture prior (Eq.~(9)) is used \emph{exclusively during training} to shape the KL objective and align the latent space with species prototypes; it imposes no label dependency at deployment time.

\subsection{KL Approximation and Tightness}
The log-sum-exp bound follows Jensen's inequality: $\mathrm{KL}\!\left(q \,\middle\|\, \sum_k \pi_k \mathcal{N}_k\right) \le \sum_k \pi_k \, \mathrm{KL}\!\left(q \,\middle\|\, \mathcal{N}_k\right)$, yielding closed-form Gaussian KL terms, the same approximation as \citep{bai2022cgmvae}. The UMAP (Figure \ref{fig:umap_vis}) shows eight well-separated semantic clusters, confirming non-collapsed posteriors and that the bound is sufficiently tight empirically.

\subsection{Gradient Analysis}
To understand the efficacy of the Imbalance-Aware Decoupled Decoding module of STELLAR, consider the gradient of the loss with respect to the output logit $z_j$. For a negative sample ($y=0$), the standard Binary Cross Entropy (BCE) gradient is simply $\frac{\partial \mathcal{L}_{\text{BCE}}}{\partial z} = p$. Even if the model predicts a low probability (e.g., $p=0.01$), the accumulated gradient from thousands of negative samples can be substantial, drowning out the sparse signal from positives.
In contrast, the gradient for ASL includes a dampening factor:
\begin{equation}
    \frac{\partial \mathcal{L}_{\text{ASL-}}}{\partial z} \approx (p_m)^{\gamma_-} \cdot p.
\end{equation}
If $p < m$, the shifted probability $p_m = 0$, and the gradient becomes exactly zero. This creates a ``hard margin'' effect, effectively silencing the noise from abundant easy negatives and redirecting model capacity toward the difficult tail species.

\subsection{Complexity Analysis.}
We provide a rigorous breakdown of the inference complexity of STELLAR, distinguishing between the shared grid-level backbone and the fine-grained per-checklist inference. 
Let $N_G$ be the number of grid cells, $N_U$ the number of survey checklists (where typically $N_U \gg N_G$), $S$ the species richness, and $d$ the latent dimension.
The computational workflow consists of three stages:

\begin{enumerate}
    \item {Grid Backbone (Shared Context):} 
    The GNN-RNN aggregates history across the spatial graph. This depends only on the grid size and is computed once per time step:
    \begin{equation}
        \mathcal{O}_{\text{grid}} = \mathcal{O}(N_G \cdot (|\mathcal{E}| + d^2)).
    \end{equation}
    Crucially, this expensive structural learning is {independent of species richness} $S$.

    \item {Local Inference (Per Checklist):} 
    For each survey $u$, we fuse the local environment $\mathbf{x}_u$ with the grid context $\mathbf{h}_c$ and pass it through the inference network (MLP) and projection head (MLP) to obtain the anchor $\tilde{\mathbf{z}}_u$. Since the input dimension is fixed (environmental features + context), this step scales with the number of surveys but is constant with respect to species:
    \begin{equation}
        \mathcal{O}_{\text{infer}} = \mathcal{O}(N_U \cdot d^2).
    \end{equation}

    \item {Geometric Decoding (Bottleneck):} 
    The final prediction is a matrix-vector product between the anchor $\tilde{\mathbf{z}}_u$ and the species prototype matrix $\mathbf{W} \in \mathbb{R}^{d \times S}$.
    \begin{equation}
        \mathcal{O}_{\text{decode}} = \mathcal{O}(N_U \cdot S \cdot d).
    \end{equation}
\end{enumerate}

The total inference cost is the sum of these stages:
\begin{equation}
    \mathcal{O}_{\text{total}} = \mathcal{O}(N_G \cdot d^2) + \mathcal{O}(N_U \cdot d^2) + \mathcal{O}(N_U \cdot S \cdot d).
\end{equation}

Although the complexity scales linearly with $N_U \cdot S$, the actual implementation benefits significantly from modern GPU tensor cores and batch processing, which enable massive parallelization of these dense operations.

\section{Visualization of Learned Species Embeddings}
\label{app:embeddings}

To understand how STELLAR models inter-species correlations, we extracted the learned species embeddings (the final representations before the classification head) for 100 randomly selected species. We projected these high-dimensional vectors into a 2D space using UMAP (Uniform Manifold Approximation and Projection) and performed K-Means clustering ($k=8$) on the projections to visualize latent structures.

\begin{figure*}[h]
    \centering
    % REPLACE 'Figures/species_umap.png' with your actual filename
    \includegraphics[width=1.0\textwidth]{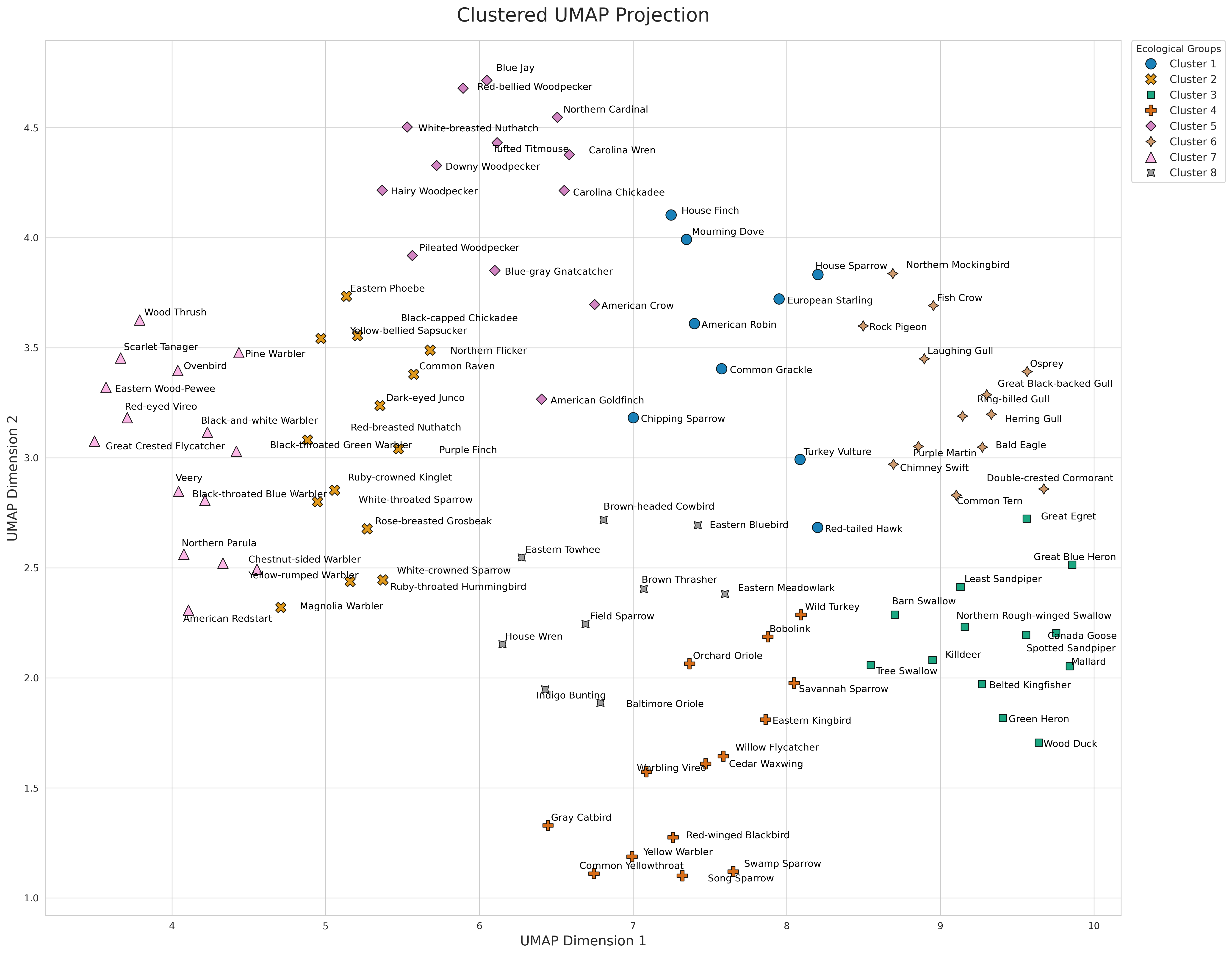}
    \caption{UMAP visualization of learned embeddings for 100 bird species. The points are colored according to clusters identified in the projected space. The clear separation of groups demonstrates STELLAR's ability to learn structured, meaningful representations of species communities, effectively grouping taxonomically or ecologically similar species together.}
    \label{fig:umap_vis}
\end{figure*}

\paragraph{Analysis of Latent Clusters.}
Figure \ref{fig:umap_vis} illustrates the resulting projection. The distinct separation of clusters indicates that STELLAR effectively captures the intrinsic co-occurrence structure of the biodiversity data. Species residing in similar habitats or exhibiting strong biological interactions are mapped to proximal regions in the latent space. For instance, the clusters align closely with ecological traits (e.g., wetland birds vs. forest dwellers), suggesting that the model goes beyond simple memorization and learns semantically meaningful representations of the ecosystem.

\section{Qualitative Visualization of Spatial Predictions}
\label{app:spatial_vis}

To substantiate the quantitative performance gains reported in the main text, we provide a complete catalog of spatial visualizations for the top-20 rarest species in the eBird dataset. Figures \ref{fig:spatial_all} through \ref{fig:spatial_all_end} compare the ground truth distributions against predictions from STELLAR and five baselines.

\paragraph{Visual Analysis.}
These visualizations confirm a consistent trend across the long-tailed distribution. Discriminative baselines (DMSE, MulSupCon) exhibit pervasive mode collapse, yielding empty prediction maps (all zeros) for the majority of these rare taxa. Continuous representation methods (SINR) tend to produce over-smoothed, diffuse probability fields. In contrast, STELLAR consistently recovers fine-grained spatial structures, accurately delineating fragmented distributions even for species with extremely limited training support.

% =================================================================
% PAGE 1 (Species 1 & 2)
% =================================================================
\begin{figure*}[p]
    \centering
    % Species 1
    \includegraphics[width=0.95\textwidth]{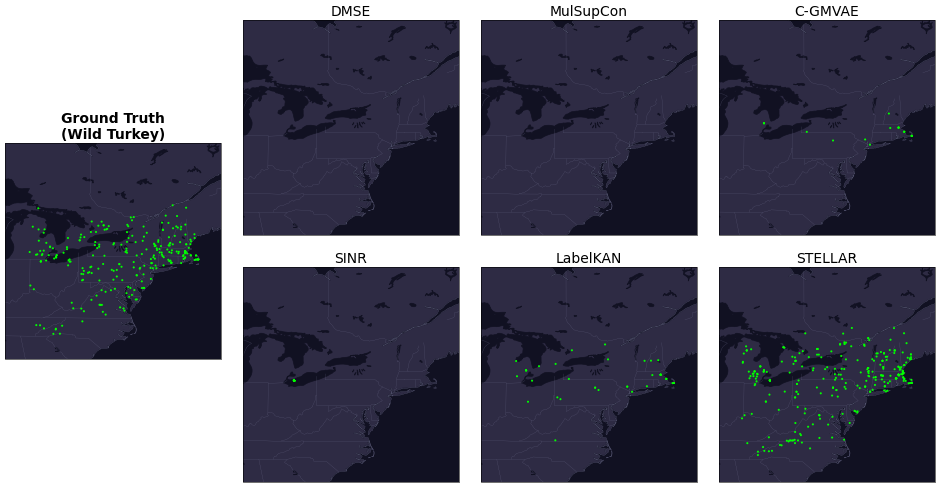}
    \vspace{5mm}
    
    % Species 2
    \includegraphics[width=0.95\textwidth]{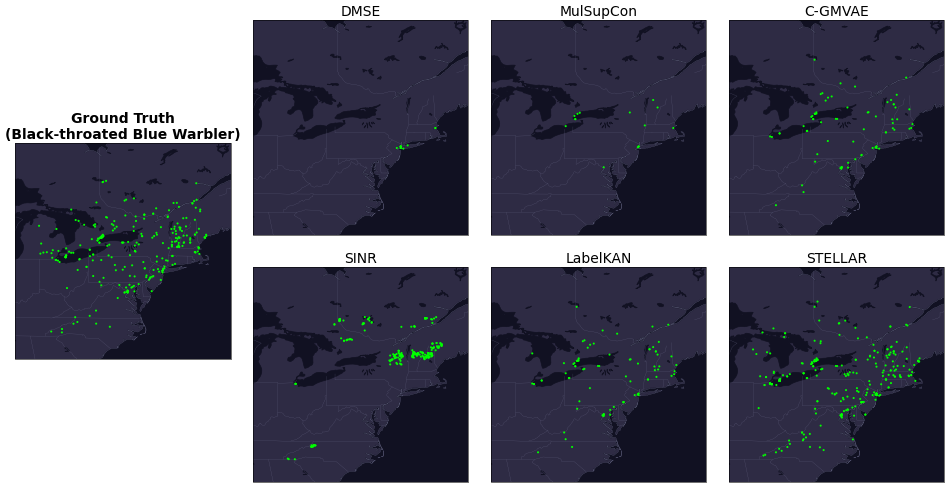}
    
    \caption{Qualitative spatial predictions for the 20 rarest species (Part 1 of 10). Rows correspond to individual species. Columns: Ground Truth, DMSE, MulSupCon, C-GMVAE, SINR, LabelKAN, and our STELLAR. STELLAR generally recovers better structure than baselines.}
    \label{fig:spatial_all}
\end{figure*}

% =================================================================
% PAGE 2 (Species 3 & 4)
% =================================================================
\begin{figure*}[p]
    \ContinuedFloat
    \centering
    % Species 3
    \includegraphics[width=0.95\textwidth]{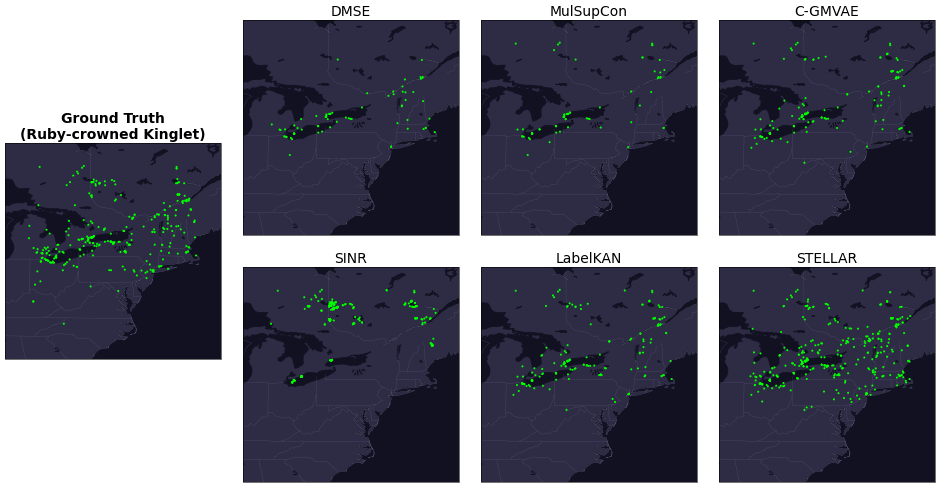}
    \vspace{5mm}
    
    % Species 4
    \includegraphics[width=0.95\textwidth]{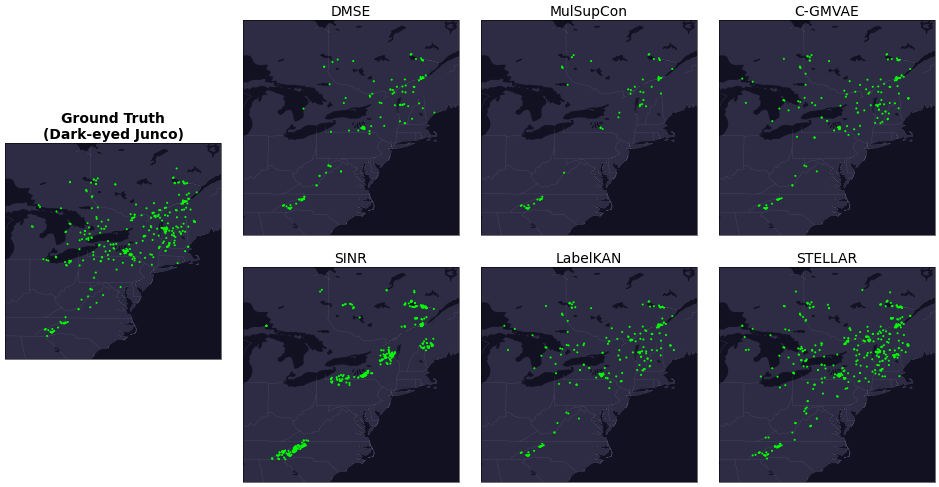}
    
    \caption[]{Qualitative spatial predictions for the 20 rarest species (Part 2 of 10).}
\end{figure*}

% =================================================================
% PAGE 3 (Species 5 & 6)
% =================================================================
\begin{figure*}[p]
    \ContinuedFloat
    \centering
    % Species 5
    \includegraphics[width=0.95\textwidth]{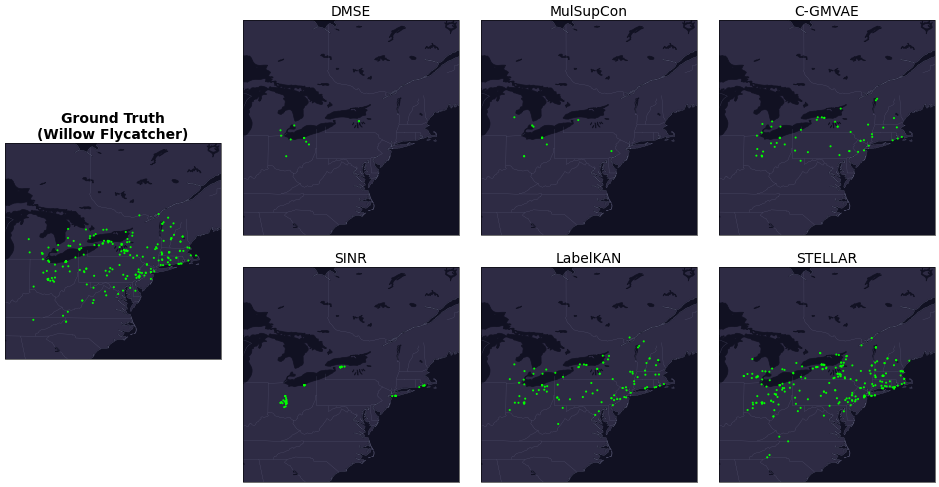}
    \vspace{5mm}
    
    % Species 6
    \includegraphics[width=0.95\textwidth]{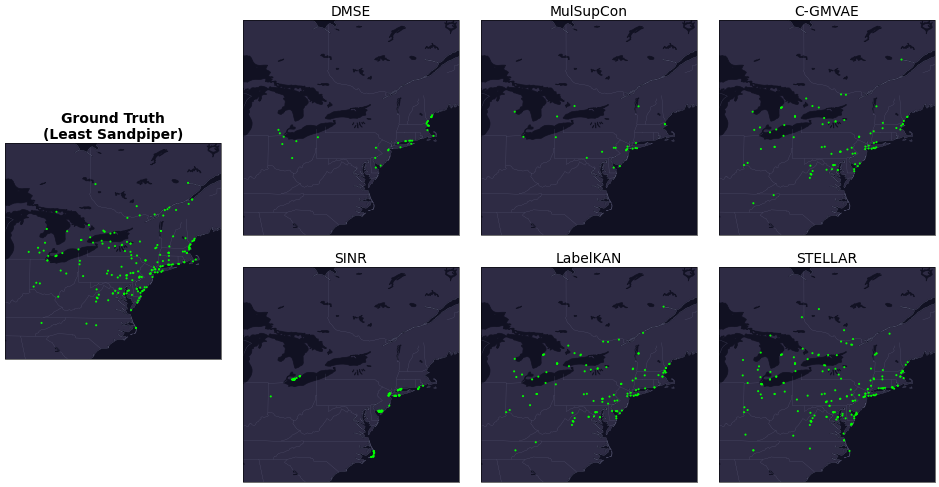}
    
    \caption[]{Qualitative spatial predictions for the 20 rarest species (Part 3 of 10).}
\end{figure*}

% =================================================================
% PAGE 4 (Species 7 & 8)
% =================================================================
\begin{figure*}[p]
    \ContinuedFloat
    \centering
    % Species 7
    \includegraphics[width=0.95\textwidth]{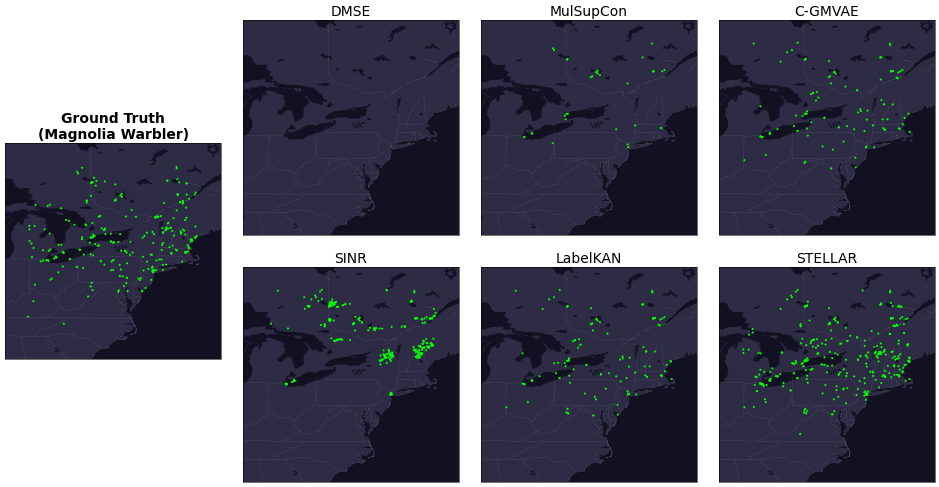}
    \vspace{5mm}
    
    % Species 8
    \includegraphics[width=0.95\textwidth]{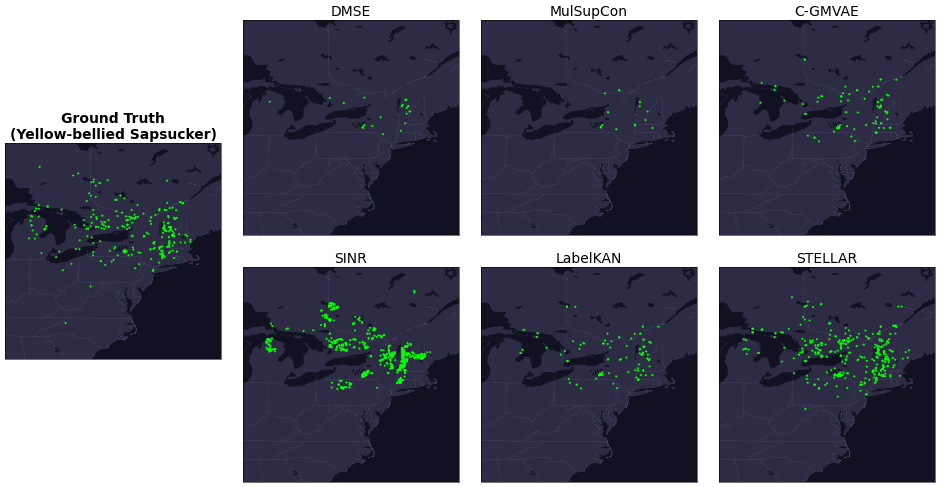}
    
    \caption[]{Qualitative spatial predictions for the 20 rarest species (Part 4 of 10).}
\end{figure*}

% =================================================================
% PAGE 5 (Species 9 & 10)
% =================================================================
\begin{figure*}[p]
    \ContinuedFloat
    \centering
    % Species 9
    \includegraphics[width=0.95\textwidth]{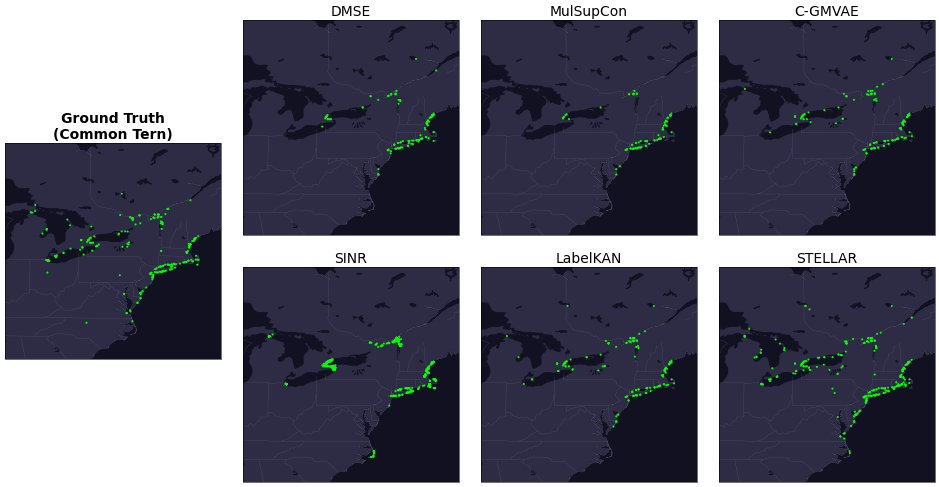}
    \vspace{5mm}
    
    % Species 10
    \includegraphics[width=0.95\textwidth]{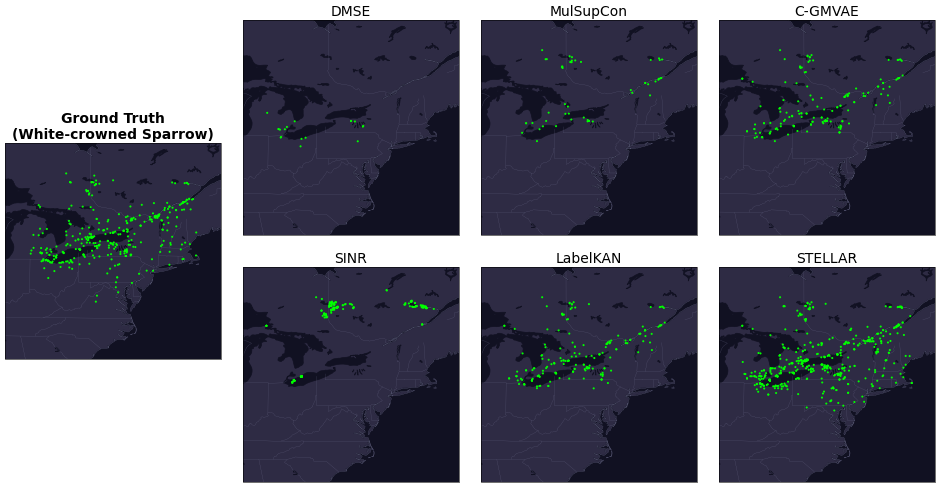}
    
    \caption[]{Qualitative spatial predictions for the 20 rarest species (Part 5 of 10).}
\end{figure*}

% =================================================================
% PAGE 6 (Species 11 & 12)
% =================================================================
\begin{figure*}[p]
    \ContinuedFloat
    \centering
    % Species 11
    \includegraphics[width=0.95\textwidth]{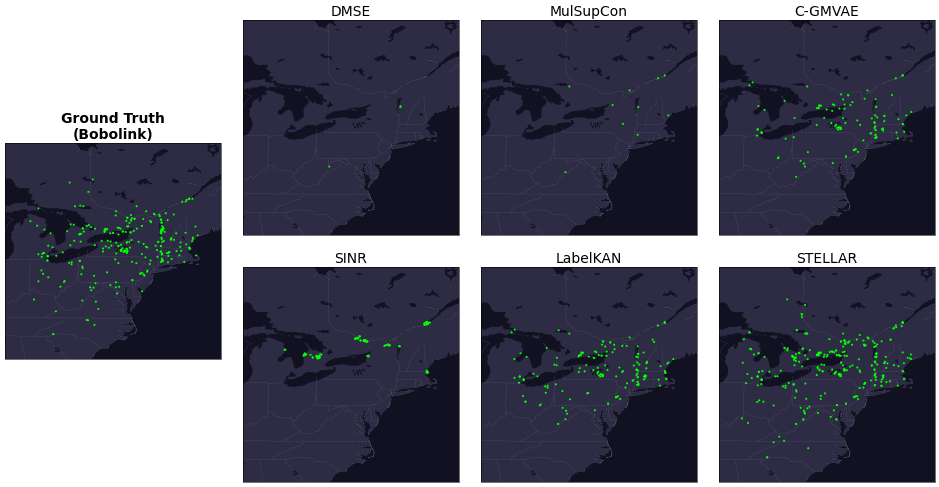}
    \vspace{5mm}
    
    % Species 12
    \includegraphics[width=0.95\textwidth]{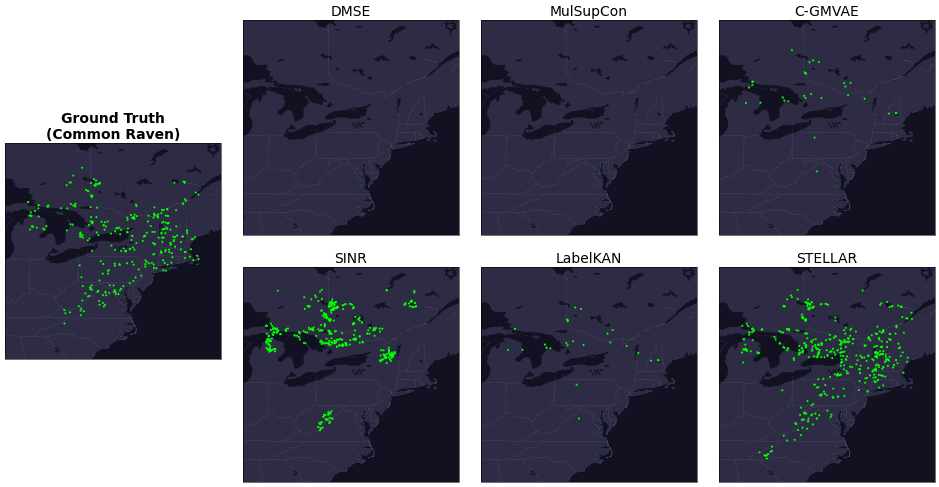}
    
    \caption[]{Qualitative spatial predictions for the 20 rarest species (Part 6 of 10).}
\end{figure*}

% =================================================================
% PAGE 7 (Species 13 & 14)
% =================================================================
\begin{figure*}[p]
    \ContinuedFloat
    \centering
    % Species 13
    \includegraphics[width=0.95\textwidth]{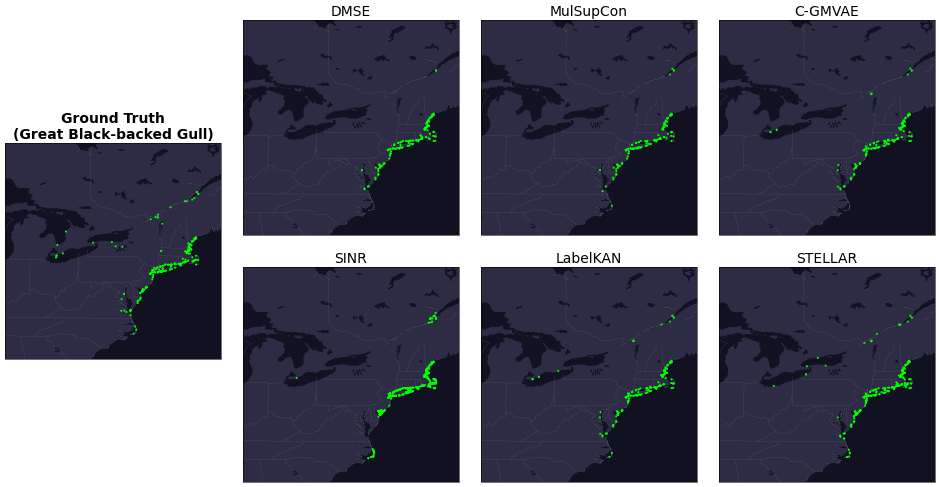}
    \vspace{5mm}
    
    % Species 14
    \includegraphics[width=0.95\textwidth]{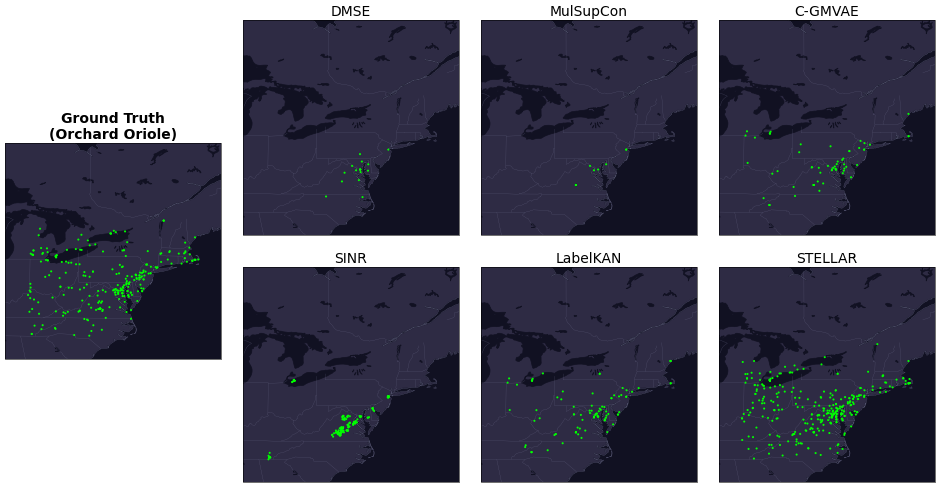}
    
    \caption[]{Qualitative spatial predictions for the 20 rarest species (Part 7 of 10).}
\end{figure*}

% =================================================================
% PAGE 8 (Species 15 & 16)
% =================================================================
\begin{figure*}[p]
    \ContinuedFloat
    \centering
    % Species 15
    \includegraphics[width=0.95\textwidth]{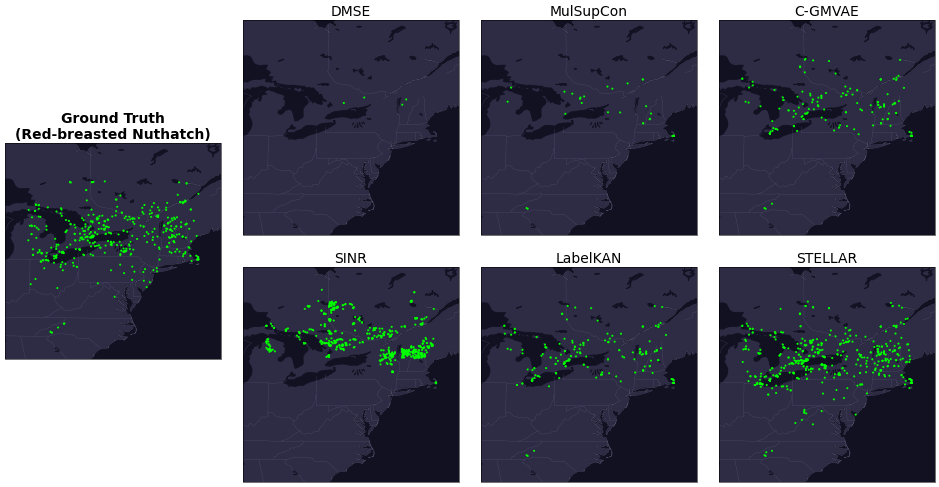}
    \vspace{5mm}
    
    % Species 16
    \includegraphics[width=0.95\textwidth]{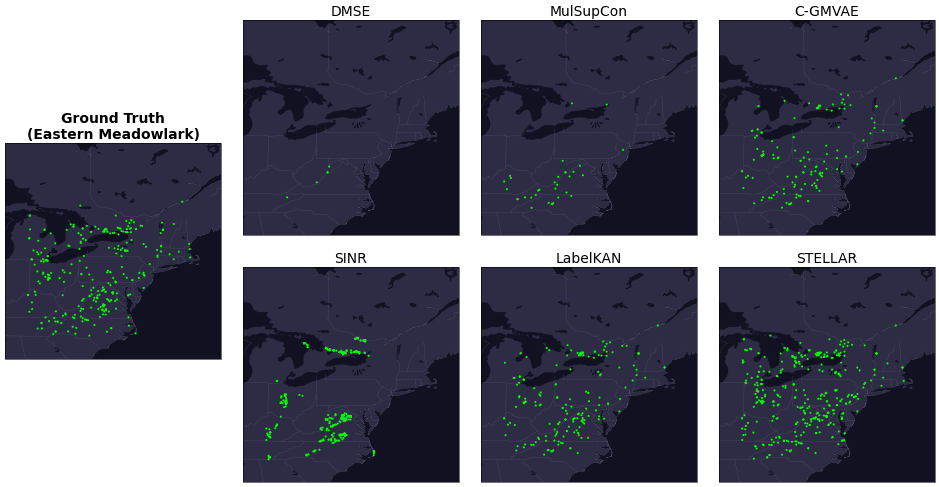}
    
    \caption[]{Qualitative spatial predictions for the 20 rarest species (Part 8 of 10).}
\end{figure*}

% =================================================================
% PAGE 9 (Species 17 & 18)
% =================================================================
\begin{figure*}[p]
    \ContinuedFloat
    \centering
    % Species 17
    \includegraphics[width=0.95\textwidth]{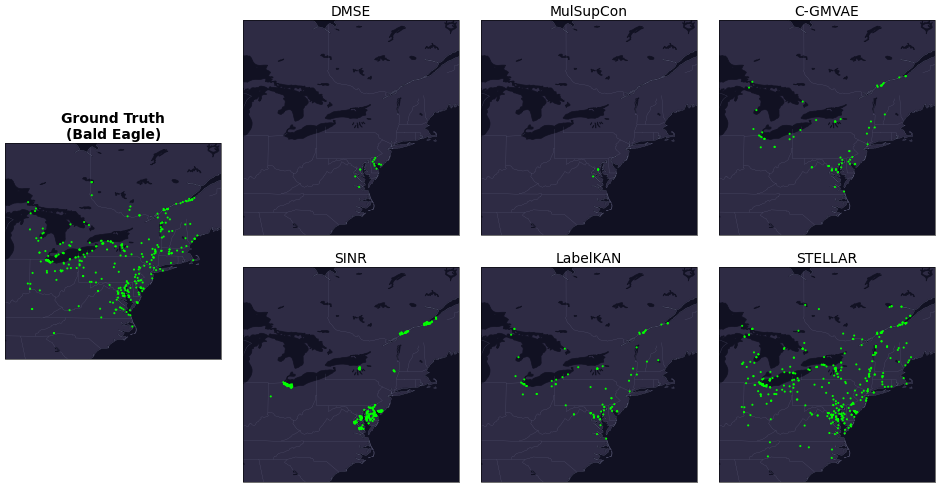}
    \vspace{5mm}
    
    % Species 18
    \includegraphics[width=0.95\textwidth]{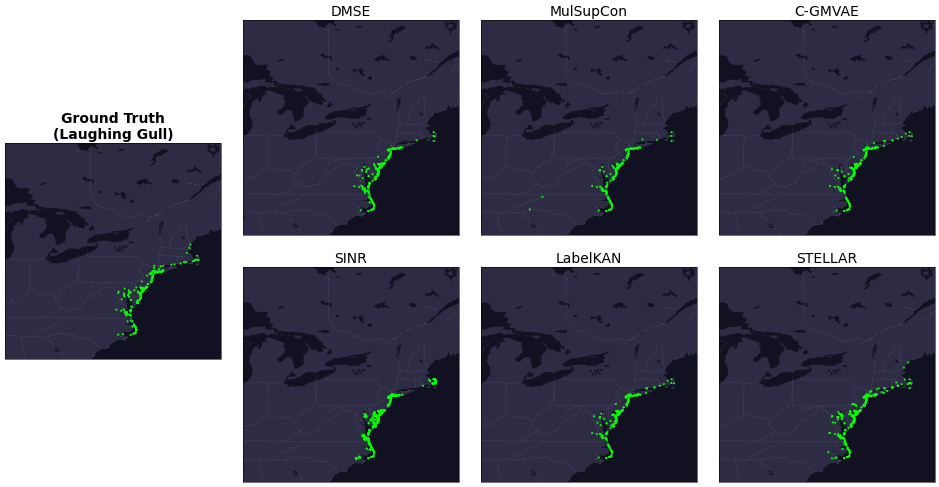}
    
    \caption[]{Qualitative spatial predictions for the 20 rarest species (Part 9 of 10).}
\end{figure*}

% =================================================================
% PAGE 10 (Species 19 & 20)
% =================================================================
\begin{figure*}[p]
    \ContinuedFloat
    \centering
    % Species 19
    \includegraphics[width=0.95\textwidth]{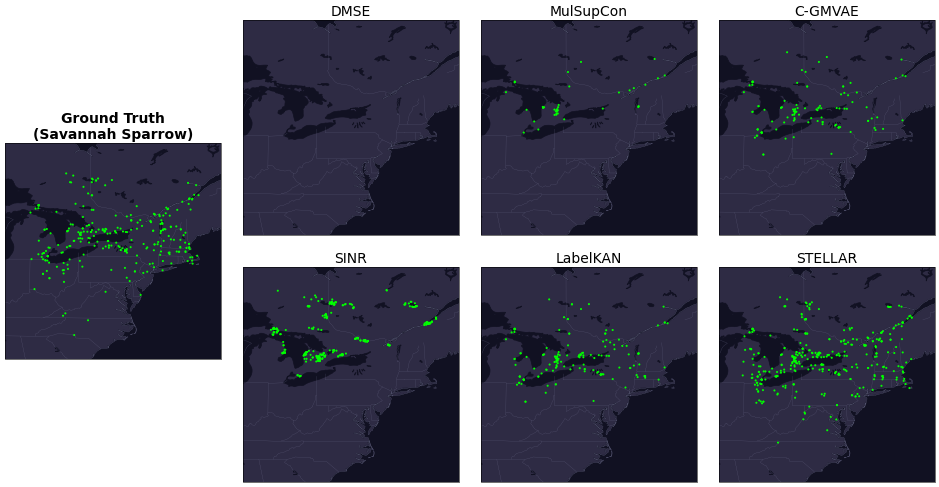}
    \vspace{5mm}
    
    % Species 20
    \includegraphics[width=0.95\textwidth]{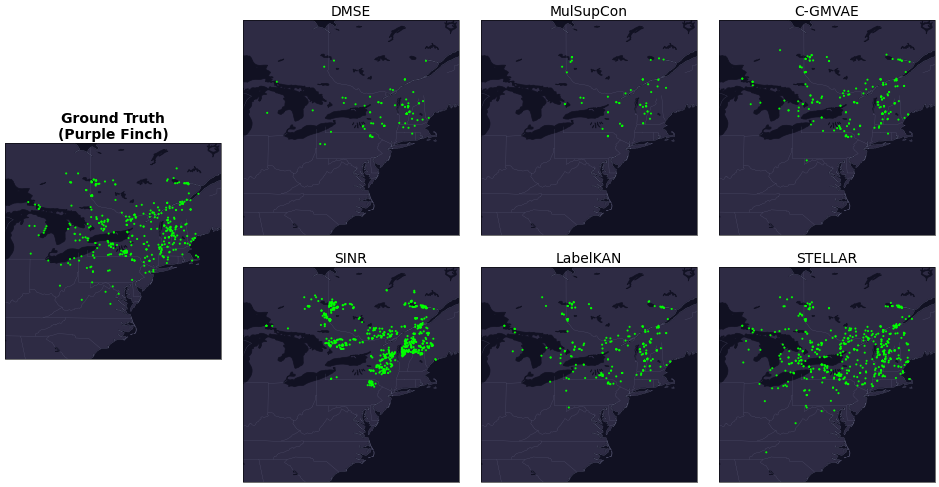}
    
    \caption[]{Qualitative spatial predictions for the 20 rarest species (Part 10 of 10).}
    \label{fig:spatial_all_end}
\end{figure*}

\end{document}